\documentclass{article}

 \usepackage[eandd,final]{neurips_2026}

\usepackage[utf8]{inputenc} 
\usepackage[T1]{fontenc}    
\usepackage{hyperref}       
\usepackage{url}            
\usepackage{booktabs}       
\usepackage{amsfonts}       
\usepackage{nicefrac}       
\usepackage{microtype}      
\usepackage{xcolor}         
\usepackage{tabularx, booktabs, multirow}
\usepackage{makecell}   

\usepackage{times}
\usepackage{caption}
\usepackage{latexsym}
\usepackage{diagbox} 
\usepackage{amsmath}   
\usepackage{amssymb}   
\usepackage{bm}        
\usepackage[T1]{fontenc}
\usepackage[utf8]{inputenc}
\usepackage{microtype}
\usepackage{inconsolata}
\usepackage{graphicx}
\usepackage[most]{tcolorbox}
\usepackage{fontawesome}
\usepackage[table]{xcolor}
\usepackage{times}
\usepackage{enumitem}
\usepackage{multirow}
\usepackage{placeins}
\usepackage{subcaption}
\usepackage{rotating} 
\usepackage{makecell}
\usepackage{algorithm}
\usepackage{algpseudocode}
\usepackage{wrapfig}
\setlist{nosep}
\usepackage{soul}
\usepackage{colortbl}   
\usepackage{pgf}  
\usepackage{multicol}

\usepackage{dblfloatfix}
\usepackage{titlesec}
\usepackage{adjustbox}
\usepackage{booktabs}
\usepackage{dblfloatfix}    
\usepackage[most]{tcolorbox}
\usepackage{microtype}
\usepackage{array}
\usepackage{tabularx}
\usepackage{booktabs}
\usepackage{adjustbox}
\usepackage{ragged2e}
\usepackage{soul}
\usepackage{longtable}
\usepackage{titlesec}

\titlespacing*{\paragraph}
{0pt}    
{0.8ex}  
{0.8ex}  

\newcommand{\boldblue}[1]{\textbf{\textcolor{blue}{#1}}}
\newcommand{\boldorange}[1]{\textbf{\textcolor{orange}{#1}}}

\newcolumntype{L}[1]{>{\raggedright\arraybackslash}p{#1}}

\newcommand{\highest}[2][50]{%
  {\setlength{\fboxsep}{0.5pt}%
  \colorbox{green!#1}{#2}}%
}

\newcommand{\lowest}[2][50]{%
  {\setlength{\fboxsep}{1.5pt}%
  \colorbox{orange!#1}{#2}}%
}
\newcommand{\highestnew}[2][30]{%
  {\setlength{\fboxsep}{1.5pt}%
  \colorbox{green!#1}{#2}}%
}

\newcommand{\NoSim}[2][30]{%
  {\setlength{\fboxsep}{1.5pt}%
  \colorbox{yellow!#1}{#2}}%
}

\title{Evaluating Whether LLMs Can Reliably Connect the DOTs?}

\author{
  Eftekhar Hossain \quad John Salvador \quad Santu Karmaker \\
  Bridge-AI Lab@UCF, Department of Computer Science \\
  University of Central Florida, USA \\
  \texttt{\{eftekhar,santu\}@ucf.edu}
}

\begin{document}

\maketitle

\begin{abstract}
Access to real-world information is often noisy and fragmented. Constructing a coherent narrative from such fragments requires models to reconstruct missing spans within a broader storyline, commonly referred to as \textit{{text infilling}}, while preserving consistency with both the local context and the global storyline. Despite using \textit{text infilling} as a pre-training objective in many Large Language Models (LLMs), their actual performance on real-world \textbf{\textit{narrative infilling}} remains underexplored. In this paper, we address this gap by introducing a multi-domain benchmark of \textbf{$\sim$9.2K} instances for \textit{narrative infilling}, constructed by masking one to three sentences across four narrative types: encyclopedic text, commonsense stories, news articles, and visual narratives. Using this benchmark, we evaluate 20 instruction-tuned open-source LLMs ranging from 1.5B to 70B parameters across varying levels of instruction specificity and reasoning guidance. Outputs are assessed using standard automatic metrics and a qualitative framework covering five narrative dimensions. Results show that model scale does not reliably predict infilling quality: \textbf{Gemma-2-2B} achieves the highest qualitative score (\textbf{4.02/5}), outperforming models over ten times larger, including \textbf{DeepSeek-Qwen-32B} (\textbf{3.77/5}, $\downarrow$6.6\%)  and \textbf{LLaMA-3.3-70B} (\textbf{3.71/5}, $\downarrow$8.3\%). We further find that explicit reasoning offers limited benefits as chain-of-thought reasoning yields only a marginal improvement (+0.6\%). Additionally,  short narratives and domain characteristics emerge as stronger predictors of task difficulty than infill position alone for narrative infilling in current LLMs. The dataset and code are available at \url{https://github.com/BridgeAI-Lab/Narrative-Infilling}.

\end{abstract}

\section{Introduction}

\begin{wrapfigure}{r}{0.52\linewidth}
\vspace{-4mm}
    \centering
    \includegraphics[width=\linewidth]{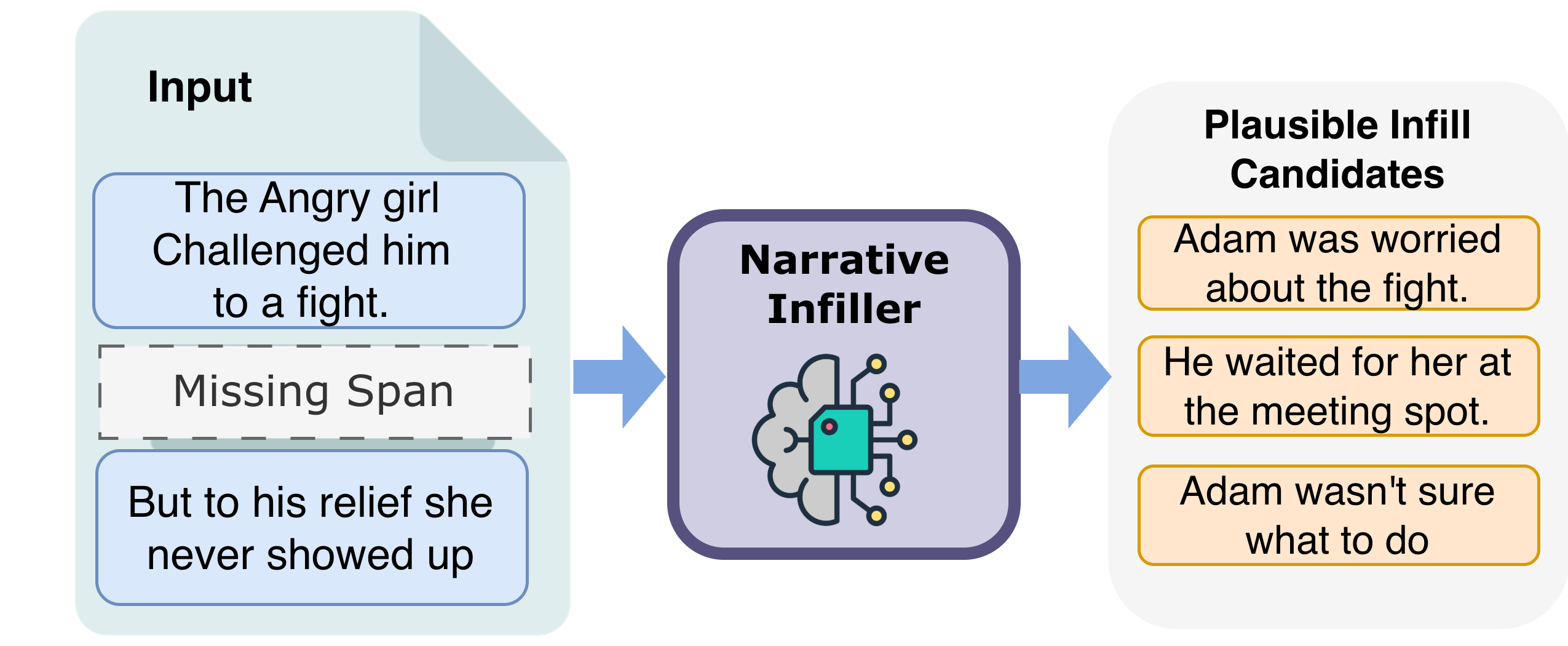}
    \caption{Narrative infilling task. Given a passage with a missing span, a \textit{Narrative Infiller} generates text that plausibly fills the gap while preserving coherence with the surrounding narrative context.}
    \vspace{-4mm}
    \label{fig:infilling}
\end{wrapfigure}

Information in real-world scenarios rarely arrives as a complete story. News reports, eyewitness accounts, and historical records often provide fragmented/partial descriptions of events, leaving important gaps in the overall narrative. Reconstructing these missing pieces is necessary to connect the fragments into a coherent account of what actually happened. While humans can perform this reconstruction effortlessly, using surrounding context and world knowledge~\citep{graesser1994constructing}, enabling machines to perform a similar process remains challenging. Reconstructing such missing information within a story is referred to as \textit{narrative infilling}. Unlike open-ended story generation~\citep{fan-etal-2018-hierarchical, rashkin2020plotmachines}, where a model continues freely from a prompt, narrative infilling requires generating a missing span that coherently bridges an existing gap while remaining consistent with both the preceding and following narrative context (Figure~\ref{fig:infilling}). Despite being widely used as a pre-training objective in modern language models~\citep{bavarian2022efficient, donahue-etal-2020-enabling}, text infilling evaluation has largely focused on constrained scenarios such as token-level completion, sentence fusion~\citep{Schwarzer_Tanprasert_Kauchak_2021, Brook_Weiss_Roit_Ernst_Dagan_2022}, or short passage bridging~\citep{Kim_Kim_Chang_Liu_2019}. These settings primarily measure local reconstruction ability, yet overlook the broader demands of narrative coherence. As a result, how well modern LLMs can reconstruct missing narrative segments in real-world narrative infilling remains an open question.


To address this gap, we construct a multi-domain narrative infilling benchmark by masking spans of 1 to 3 consecutive sentences from narratives drawn from several real-world sources, including encyclopedic text (Wikipedia), commonsense stories (ROCStories), news articles (CNN/DailyMail), and visual narratives (SIND). Using this benchmark, we evaluate 20 instruction-tuned open-source LLMs to enable reproducible, accessible comparisons. In particular, we analyze two factors that may influence infilling quality: \textit{prompt/instruction specificity} and \textit{explicit reasoning guidance}, and investigate how narrative type, length, and infill position affect task difficulty.



Our results reveal several insights about narrative infilling in LLMs. First, instruction specificity plays an important role: increasing instruction detail from minimal to moderate improves automatic scores by \textbf{16.7\%} and qualitative scores by \textbf{3\%}, while additional detail yields little further improvement. Second, reasoning guidance yields uneven effects: chain-of-thought reasoning provides only a marginal improvement (+0.6\%) over the moderate baseline. Overall, coherent narrative infilling remains challenging for current models, with a mean qualitative score of \textbf{3.80/5} across all models. Even the strongest models achieve a score of \textbf{4.02/5}. Performance also varies substantially across model families and instruction strategies, with scores ranging from \textbf{2.33} to \textbf{4.02} (std $\approx$0.16). Together, these findings provide new insights into the capabilities and limitations of large language models for narrative infilling. Our contributions are threefold:


\begin{itemize}[leftmargin=*, itemsep=1pt, topsep=2pt]

\item We introduce a multi-domain benchmark for narrative infilling containing \textbf{9,142} instances with controlled variation in span length and blank position, enabling systematic analysis of structural factors that affect infilling difficulty. 


\item We developed a qualitative evaluation framework for narrative infilling that assesses five narrative dimensions and applies it alongside standard 
automatic metrics to systematically compare 20 LLMs (1.5B–70B parameters) under varying levels of instructions and reasoning strategies.


\item We identify several key insights about narrative infilling in LLMs: 
(1) moderate increases in instruction specificity improve automatic and qualitative performance by $16.7\%$ and $3\%$, respectively;  
(2) explicit reasoning guidance gives only marginal improvements (<$1\%$); and (3) infilling difficulty increases for shorter narratives and when the missing span occurs at narrative boundaries (opening/closing positions). 
\end{itemize}

\section{Related Work}

\noindent\textbf{Narrative Completion and Infilling.}
Narrative completion has long been used to study story understanding and commonsense reasoning in language models. Early cloze-style benchmarks such as the Story Cloze Test~\citep{srinivasan2018simple} and ROCStories~\citep{mostafazadeh-etal-2016-corpus} evaluate models' ability to select coherent story endings, while subsequent work explores story continuation and controllable generation conditioned on plot structures 
or commonsense knowledge~\citep{fan-etal-2018-hierarchical, 
rashkin2020plotmachines, goldfarb2020content, ammanabrolu2021automated}. 
A related line of work studies text infilling, proposing specialized architectures and training objectives for generating missing spans~\citep{donahue-etal-2020-enabling, shen2020blank, bavarian2022efficient, du2022glm}. One work by \citet{paul2021coins} begins to explore narrative-level infilling, but remains limited to customized models and predominantly single-direction reasoning. As a result, existing approaches largely focus on local completion or continuation rather than reconstructing missing segments within broader narrative contexts. Consequently, the ability of LLMs to reconstruct missing narrative segments while maintaining global narrative coherence remains largely unexplored. We address this gap by introducing a narrative infilling benchmark and systematically evaluating 20 LLMs using both automatic metrics and qualitative judgments.

\noindent\textbf{Instruction Sensitivity and Reasoning in LLMs.}
Recent work shows that LLM behavior can vary depending on how tasks are specified through instructions or prompts \citep{zhuo2024prosa, chatterjee2024posix}. Accordingly, researchers have explored whether explicit reasoning guidance—such as chain-of-thought reasoning~\citep{wei2022chain} and its variants~\citep{Wang2022SelfConsistencyIC, yao2023tree} can improve performance on reasoning-intensive tasks. Related research also studies reasoning in narrative and commonsense settings, including abductive reasoning for explaining incomplete narratives \citep{bhagavatulaabductive, zhao2024uncommonsense}, causal reasoning for modeling event relationships in stories \citep{ammanabrolu2021automated, mostafazadeh2020glucose, romanou2023crab}, and counterfactual reasoning for revising narratives under hypothetical changes \citep{qin2019counterfactual, qin2020back, wu2021polyjuice}. Despite these advances, the role of reasoning strategies and instruction design in reconstructing missing narrative segments across diverse genres remains largely unexplored. 

\section{Method}

\subsection{Dataset Construction}
To evaluate narrative infilling across a broad spectrum of text genres, we constructed a benchmark dataset by curating samples from three publicly available story corpora: \textbf{SIND} \citep{huang2016visual}, \textbf{ROCStories} \citep{mostafazadeh-etal-2016-corpus}, and \textbf{CNN/DailyMail} \citep{see-etal-2017-get}, along with
\textbf{Wikipedia}\footnote{\url{https://dumps.wikimedia.org/}} articles as a source of factual prose. These sources were selected to capture a diverse range of narrative forms and discourse structures, including factual prose, visual storytelling, short fictional narratives, and long-form journalistic texts. This diversity allows us to probe model generalization across both tightly structured and loosely organized narrative forms.

\begin{wraptable}{r}{0.52\linewidth}
\vspace{-5mm}
\caption{Dataset statistics: number of samples and average number of sentences per sample from the datasets.}
\tiny
\centering
\begin{adjustbox}{width=\linewidth,center}
\begin{tabular}{l|ccc}
\toprule
\textbf{Dataset} & \textbf{\#Samples} & \textbf{Avg. Sent.} & \textbf{\#FIB Problems}\\
\midrule
ROCStories       & 1,000 & 5      & 3,000 \\
SIND             & 1,000 & 5      & 1,182 \\
CNN/DailyMail       & 975   & 33.37  & 2,728 \\
Wikipedia          & 848   & 39.57  & 2,232 \\
\midrule
Total                & 3,823 & -      & 9,142 \\
\bottomrule
\end{tabular}
\end{adjustbox}

\label{tbl:dataset_stats}

\end{wraptable}

\smallskip\noindent\textbf{Sampling and Masking.}
From each corpus, we randomly sampled approximately 1,000 entries, yielding a combined pool of \textbf{3,823} base samples. To construct infilling instances, we masked contiguous spans of one, two, or three sentences by replacing them with a placeholder token. Spans were extracted at varying positions within each narrative to ensure coverage of different discourse roles, including opening, middle, and closing segments. Shorter spans of one or two sentences will test the model's ability to infer locally coherent content from the immediate context, whereas a span of three sentences demands more substantive reasoning about narrative structure and discourse flow. We capped the maximum span at three sentences to ensure that the surrounding context remains sufficient for meaningful reconstruction because longer masks may risk removing too much narrative information, making the task ill-defined rather than genuinely challenging. Applying this procedure across all span lengths and positions yielded \textbf{9,142} unique narrative infilling instances, distributed across the four source domains. Table~\ref{tbl:dataset_stats} provides a statistical overview of the final dataset. 
Regarding the position of the missing spans, the dataset contains opening (5,314), middle (2,700), and closing (1,128) samples.




\subsection{Large Language Models Investigated}
\label{sec:llms}
We benchmark 20 open source LLMs drawn from six prominent model families: Gemma~\citep{gemma_2024}, LLaMA~\citep{grattafiori2024llama3herdmodels}, Mistral~\citep{mistral}, Qwen~\citep{qwen25}, DeepSeek-R1~\citep{deepseekai2025deepseekr1incentivizingreasoningcapability}, and OLMo~\citep{olmo20252olmo2furious} covering a broad range of parameter
scales. Specifically, the models include Gemma-2 (2B, 9B, 27B); LLaMA-3.1-8B, LLaMA-3.2 (1B, 3B), and LLaMA-3.3-70B; Mistral (7B, 24B), and Ministral-8B; Qwen-2.5 (7B, 14B); DeepSeek-R1-Distill-Qwen (1.5B, 7B, 14B, 32B), DeepSeek-R1-Distill-LLaMA (8B, 70B); Olmo-2 (7B, 13B). All models are queried under controlled decoding parameters: temperature $= 0.8$, top-$p = 0.9$, and a maximum output length set to twice the length of the masked span. These settings promote sufficient generative variability while preserving coherent and contextually grounded outputs. 

\subsection{Instruction Design}
To probe how models respond to varying levels of task instruction, we structure our study along two complementary axes: \textit{instruction specificity} and \textit{reasoning guidance}. All the instruction templates are provided in the Appendix~\ref{sec:prompt-template}.

\noindent\textbf{Instruction specificity. }
To evaluate how sensitive models are to the level of instruction detail, 
we adopt the TELeR (Turn, Expression, Level of Details, Role) taxonomy \citep{santu2023teler}, which provides a principled way to vary task presentation across various levels of increasing explicitness. We create prompts corresponding to TELeR Levels 0 (very low specificity) through 4 (very high specificity). The five levels are as follows:


\begin{itemize}[leftmargin=*, nosep, topsep=2pt]
    \item \textbf{Very Low Specificity (Level 0):}  A minimal instruction that presents only the raw input to the model without any contextual guidance.
    \item \textbf{Low Specificity (Level 1):} A high-level directive that instructs the model to generate a narrative infill without any further task-specific guidance.
    \item \textbf{Moderate Specificity (Level 2):} A multi-sentence paragraph that articulates the task through several complementary sub-tasks.
    \item \textbf{High Specificity (Level 3):} A bulleted list in which each item describes a distinct sub-task, providing structured and explicit instructions.
    \item \textbf{Very High Specificity (Level 4):} Extends Level 3 by incorporating few-shot examples to guide the model for infilling.
\end{itemize}
\noindent\textbf{Reasoning Strategies. } 
We design a second set of instructions by pairing the moderate specificity (Level 2) with one of four reasoning paradigms. The paradigms are as follows: 



\begin{itemize}[leftmargin=*, nosep, topsep=2pt]
\item \textbf{Abductive Reasoning:} The model infers the most plausible explanation that connects the surrounding narrative context 
(e.g., \textit{“Infer the most likely event that explains the missing part of the story given the surrounding context.”}).

\item \textbf{Causal Reasoning:} The model generates a fill that preserves cause–and–effect relationships within the narrative 
(e.g., \textit{“Identify the causal relationships in the surrounding narrative and generate a sentence that maintains this logic.”}).

\item \textbf{Chain-of-Thought Reasoning:} The model is prompted to reason step-by-step before producing the infilled span 
(e.g., \textit{“First reason step-by-step about what likely happened in the story, then generate the missing text.”}).

\item \textbf{Counterfactual Reasoning:} The model considers alternative possibilities before generating the infill 
(e.g., \textit{“Consider possible alternative events consistent with the story and generate the most plausible missing part.”}).
\end{itemize}

\subsection{Evaluation}
\label{sec:evaluation}
\noindent\textbf{Automatic Evaluation.}
We employed five widely used automatic evaluation metrics: BERTScore~\citep{bertscore}, ROUGE-1~\citep{rouge}, 
chrF~\citep{chrf}, METEOR~\citep{meteor} and Sem-F1~\citep{semf1}. 
While these metrics are efficient and widely used to evaluate generated outputs, they primarily capture surface-level similarity and do not fully capture the narrative qualities required for coherent text infilling. 

\noindent\textbf{Infilling Quality Evaluation. }
To better capture narrative-level properties, we design a five-dimensional evaluation rubric that focuses on aspects most critical to narrative infilling. Each dimension is rated on a five-point Likert scale:

\begin{itemize}[leftmargin=*, nosep, topsep=4pt]
    \item \textbf{Fluency:} Measures the naturalness and grammatical correctness of the generated text.
    \item \textbf{Context Faithfulness:} How well the completion remains grounded in the surrounding narrative without introducing unsupported claims.
    \item \textbf{Bidirectional Coherence:} Measures how smoothly the completion connects the preceding and following context.
    \item \textbf{Narrative Consistency:} Measures whether the completion preserves story elements such as timeline, characters, tone, and causal structure.
    \item \textbf{Informativeness:} How effectively the completion fills the narrative gap without being generic.
\end{itemize}

\noindent\textbf{LLM-as-Judge Evaluation.}
Although these rubrics capture important narrative qualities, manually evaluating each dimension across all model outputs (20 LLMs, five prompt levels, and four reasoning paradigms) would be prohibitively expensive and time-consuming. Recent studies have shown that strong LLMs can serve as reliable evaluators for generated text~\citep{li2024llms, zheng2023judging, wang2023chatgpt}. Motivated by this line of work, we adopt an \textit{LLM-as-Judge} approach to enable scalable evaluation. 
Specifically, we use \texttt{GPT-4.1-mini} (version 2025-04-14) to score each completion across the five rubric dimensions. To ensure deterministic evaluation, we set the judge temperature to 0.


\noindent\textbf{Human Evaluation on a Subset.}
To assess the reliability of \texttt{GPT-4.1-mini} as an evaluator, we conducted a human validation study on a randomly sampled subset of 300 completions. Each sample was independently annotated by two human evaluators using the same rubric. The annotators have more than 4 years of research experience in NLP and AI, as well as prior experience with annotation and qualitative evaluation of generation tasks. Each annotator spent an average of eight minutes per sample, carefully reading the surrounding narrative context before assigning scores.

\begin{wraptable}{r}{0.48\linewidth}
\vspace{-4mm}
\caption{Human--human agreement (Cohen’s $\kappa$) and human--GPT correlation (Spearman’s $\rho$) across five infilling quality dimensions.}
\centering
\scriptsize
\resizebox{\linewidth}{!}{%
\begin{tabular}{lccc}
\toprule
 Dimensions & $A_1$ vs $A_2$ & GPT vs $A_1$ & GPT vs $A_2$\\
\midrule
Fluency  &  0.701 &  0.637  & 0.580 \\
Faithfulness &  0.704 & 0.545 & 0.539 \\
Coherence & 0.717 & 0.593  & 0.549 \\
Narrative Consistency & 0.490 & 0.648 & 0.571 \\
Informativeness  & 0.656 & 0.582  & 0.579 \\
\midrule
\textbf{Average} & \textbf{0.654} & \textbf{0.601} & \textbf{0.564}\\
\bottomrule
\end{tabular}
}
\label{tab:human-vs-gpt}
\vspace{-3mm}
\end{wraptable}

We measure inter-annotator agreement (reported in Table~\ref{tab:human-vs-gpt}) using Cohen’s $\kappa$, and observe moderate overall agreement (average $\kappa=0.654$) between human annotators. Agreement is highest for Coherence ($0.717$) and Faithfulness ($0.704$) dimensions, which have relatively clear surface-level signals, while Narrative Consistency yields the lowest agreement ($0.490$), likely reflecting its dependence on deeper, global interpretation of the narrative. To assess alignment with the automatic evaluator, we compute Spearman’s rank correlation between human and GPT-based scores. As shown in Table~\ref{tab:human-vs-gpt}, correlation scores are also moderate, averaging $0.601$ and $0.564$ with annotators $A_1$ and $A_2$, respectively. Notably, the gap between human–human and human–GPT agreement remains small, suggesting that much of the discrepancy arises from the inherent subjectivity of narrative infilling evaluation rather than systematic limitations of \texttt{GPT-4.1-mini}. These agreement levels fall within commonly accepted ranges for validating LLM-based evaluators in prior work~\citep{li2024leveraging, liu2023g}, supporting the use of \texttt{GPT-4.1-mini} as a reliable and scalable proxy for large-scale evaluation.




\section{Results and Analysis} 

\subsection{Effect of Prompt Specificity} 

\begin{table}[]

\caption{Average scores across five instruction specificity levels for all 20 models. AtAvg is the mean of five automatic metrics (BERTScore, ROUGE-1, METEOR, chrF, Sem-F1); QAvg is the mean across five qualitative dimensions ( Faithfulness, Coherence, Narrative Consistency, Fluency, and Informativeness). \colorbox{green!30}{Green} indicates the highest score overall; \textbf{\textcolor{blue}{bold blue}} and \textbf{\textcolor{orange}{orange}} indicate the highest and lowest scores per level.}
\vspace{1mm}
\centering
\small
\setlength{\tabcolsep}{3pt}
\resizebox{0.9\linewidth}{!}{
\begin{tabular}{l cc cc cc cc cc}
\toprule
& \multicolumn{2}{c}{\textbf{Very Low (L0)}} 
& \multicolumn{2}{c}{\textbf{Low (L1)}} 
& \multicolumn{2}{c}{\textbf{Moderate (L2)}} 
& \multicolumn{2}{c}{\textbf{High (L3)}} 
& \multicolumn{2}{c}{\textbf{Very High (L4)}} \\
\cmidrule(lr){2-3}\cmidrule(lr){4-5}\cmidrule(lr){6-7}\cmidrule(lr){8-9}\cmidrule(lr){10-11}
Model & AtAvg & QAvg & AtAvg & QAvg & AtAvg & QAvg & AtAvg & QAvg & AtAvg & QAvg \\
\midrule
LLaMA-3.1-8B      & \boldblue{0.311} & \boldblue{3.878} & 0.354 & 3.831 & 0.357 & 3.933 & 0.352 & \boldblue{4.018} & 0.306 & 3.777 \\
LLaMA-3.2-1B      & 0.309 & 3.792 & \boldorange{0.299} & 3.534 & \boldorange{0.246} & 3.607 & \boldorange{0.256} & 3.617 & 0.240 & 3.502 \\
LLaMA-3.2-3B      & 0.304 & 3.820 & 0.349 & 3.583 & 0.335 & 3.824 & 0.305 & 3.783 & 0.249 & 3.462 \\
LLaMA-3.3-70B     & 0.314 & 3.707 & 0.375 & 3.642 & 0.375 & 3.647 & 0.371 & 3.647 & 0.258 & 3.367 \\
Ministral-8B      & 0.300 & 3.788 & 0.355 & 3.811 & 0.352 & 3.893 & 0.346 & 3.922 & 0.311 & \boldblue{3.911} \\
Mistral-nemo      & 0.287 & 3.754 & 0.352 & 3.807 & 0.357 & 3.846 & 0.354 & 3.914 & 0.282 & 3.825 \\
Mistral-small     & 0.307 & 3.652 & \highestnew{0.376} & 3.694 & 0.371 & 3.719 & 0.365 & 3.730 & 0.357 & 3.814 \\
Gemma-2-2B        & 0.286 & 3.746 & 0.342 & \highestnew{4.021} & 0.337 & \boldblue{3.988} & 0.338 & 4.003 & 0.284 & 3.574 \\
Gemma-2-9B        & 0.300 & 3.760 & 0.358 & 3.825 & 0.358 & 3.862 & 0.357 & 3.873 & 0.324 & 3.834 \\
Gemma-2-27B       & 0.311 & 3.754 & 0.365 & 3.794 & 0.360 & 3.791 & 0.359 & 3.786 & 0.351 & 3.864 \\
Olmo-2-7B         & 0.285 & 3.680 & 0.343 & 3.862 & 0.329 & 3.904 & 0.334 & 3.906 & 0.295 & 3.881 \\
Olmo-2-13B        & 0.260 & 3.639 & 0.346 & 3.802 & 0.346 & 3.919 & 0.348 & 3.905 & 0.251 & 3.643 \\
Qwen-2.5-7B       & 0.310 & 3.822 & 0.352 & 3.715 & 0.350 & 3.762 & 0.347 & 3.767 & 0.306 & 3.695 \\
Qwen-2.5-14B      & \boldblue{0.332} & 3.818 & 0.365 & 3.781 & 0.359 & 3.740 & 0.359 & 3.754 & 0.312 & 3.725 \\
DeepSeek-LLaMA-8B & 0.276 & 3.543 & 0.335 & 3.825 & 0.341 & 3.823 & 0.326 & 3.763 & 0.257 & 3.394 \\
DeepSeek-LLaMA-70B& 0.291 & 3.627 & 0.346 & 3.865 & 0.353 & 3.862 & 0.352 & 3.859 & 0.337 & 3.777 \\
DeepSeek-Qwen-1.5B& 0.268 & \boldorange{3.237} & 0.315 & \boldorange{3.293} & 0.313 & \boldorange{3.288} & 0.304 & \boldorange{3.316} & \boldorange{0.229} & \boldorange{2.328} \\
DeepSeek-Qwen-7B  & \boldorange{0.264} & 3.496 & 0.334 & 3.665 & 0.331 & 3.666 & 0.325 & 3.611 & 0.251 & 3.335 \\
DeepSeek-Qwen-14B & 0.318 & 3.764 & 0.365 & 3.885 & 0.369 & 3.878 & 0.369 & 3.885 & 0.335 & 3.792 \\
DeepSeek-Qwen-32B & 0.327 & 3.672 & 0.374 & 3.762 & \boldblue{0.375} & 3.763 & \boldblue{0.375} & 3.774 & \boldblue{0.353} & 3.715 \\
\midrule
\midrule
\rowcolor{gray!20} Overall & 0.298 & 3.700 & 0.345 & 3.762 & \textbf{0.348} & \textbf{3.800} & 0.344 & 3.781 & 0.295 & 3.643 \\
\bottomrule
\end{tabular}
}
\vspace{-3mm}
\label{tab:fitb_main}

\end{table}

Table~\ref{tab:fitb_main} reports average automatic and qualitative scores for all 20 models across the five specificity levels; per-metric breakdowns are provided in the Appendix (Tables~\ref{tab:fitb_full1}  and ~\ref{tab:fitb_full2}) to preserve space for deeper analysis of the key findings. We highlight three key findings below:

\noindent\textbf{Prompt Specificity Helps, Until It Does Not.} 
Moving from very low prompt specificity (L0) to moderate specificity (L2) yields the most consistent improvements: the average qualitative score increases from $3.70$ to $3.80$ ($+2\%$) and the average automatic score from $0.298$ to $0.348$ ($+16.7\%$), indicating that even minimal task instructions meaningfully guide model behavior. However, performance plateaus at high specificity (L3) and drops sharply at very high specificity (L4), with the average qualitative score falling to $3.643$ ($-3.7\%$ relative to L3). This decline is particularly pronounced in several models; for instance, \textit{DeepSeek-Qwen-1.5B} falls to a QAvg of $2.328$, suggesting that for this task, few-shot demonstrations shift model attention away from the surrounding narrative context, leading to less contextually grounded generations.

\noindent\textbf{Automatic and Qualitative Metrics Tell Different Stories.}
Performance trends also reveal a clear mismatch between automatic and qualitative evaluations. Models that lead on AtAvg do not consistently achieve the highest QAvg on qualitative dimensions (Table~\ref{tab:fitb_main}). For example, \textit{Gemma-2-2B} attains the highest QAvg at L1 ($4.021$) and L2 ($3.988$), despite its automatic scores trailing those of larger models such as \textit{DeepSeek-Qwen-32B} and \textit{Mistral-small (24B)}, which dominate on lexical overlap metrics. This discrepancy is further reflected in the system-level Kendall $\tau$ correlations reported in Table~\ref{tab:teler_corr_cross}. Across the five qualitative dimensions, automatic metrics correlate only weakly with judgments of those dimensions. The strongest observed correlation is just $0.141$ (between chrF and Informativeness), underscoring how poorly surface-level overlap captures the subtleties of narrative quality. For more details on the system-level correlation computation, see Appendix~\ref{app:system-level-corr}. 

\noindent\textbf{Scale Does Not Guarantee Quality.}
Finally, increasing model scale does not necessarily lead to stronger performance. For example, the 2B-parameter \textit{Gemma-2-2B} achieves the highest average qualitative score of $4.021$ (Table~\ref{tab:fitb_main}), outperforming much larger models such as  \textit{DeepSeek-Qwen-32B} ($3.774$, L3) and \textit{LLaMA-3.3-70B} ($3.707$, L0). This corresponds to improvements of roughly $+6.5\%$ and $+8.5\%$, respectively. These results suggest that narrative infilling places greater emphasis on context sensitivity and instruction following than on raw model scale. While larger models often produce fluent and globally plausible text, their generations can still drift from the intended narrative flow, reducing coherence, faithfulness, and narrative consistency.

\begin{table}
\vspace{1mm}
\caption{System-level Kendall $\tau$ correlation between narrative quality evaluations and automatic metrics computed by following \citet{revisiting_gold}. The highest correlation is shown in green.}
\vspace{2mm}
\centering
\small
\resizebox{0.7\linewidth}{!}{%
\begin{tabular}{lccccc}
\toprule
 Dimensions & BERT & ROUGE-1 & METEOR & chrF & Sem-F1 \\
\midrule
Fluency  & 0.044 & 0.074 & 0.090 & 0.085 & 0.090 \\
Faithfulness & 0.021 & 0.060 & 0.085 & 0.083 & 0.076 \\
Coherence & 0.035 & 0.068 & 0.093 & 0.091 & 0.084 \\
Narrative Consistency & 0.054 & 0.064 & 0.089 & 0.082 & 0.084 \\
Informativeness  & 0.045 & 0.099 & \highestnew[25]{0.136} & \highestnew[50]{0.141} & \highestnew[15]{0.109} \\
\bottomrule
\end{tabular}
}

\label{tab:teler_corr_cross}
\end{table}

\subsection{Effect of Reasoning Strategies} 
Table~\ref{tab:reasoning_main} provides the average evaluation scores for four reasoning strategies. In our preliminary experiments, we found that models with $<7B$ parameters produced degenerate outputs when given explicit reasoning instructions; therefore, we focused on evaluating the 10 largest models. Detailed results across each metric are reported in the Appendix (Table~\ref{tab:reasoning_full}).

\begin{table}[]

\caption{Results across four reasoning paradigms for the 10 largest models, evaluated at moderate specificity (L2). \colorbox{green!30}{Green} indicates the highest score overall; \textbf{\textcolor{blue}{bold blue}} and \textbf{\textcolor{orange}{orange}} indicate the highest and lowest per paradigm. }
\vspace{1mm}
\centering
\small
\setlength{\tabcolsep}{3pt}
\resizebox{0.8\linewidth}{!}{
\begin{tabular}{l cc cc cc cc}
\toprule
& \multicolumn{2}{c}{\textbf{Abductive}} 
& \multicolumn{2}{c}{\textbf{Causal}} 
& \multicolumn{2}{c}{\textbf{Chain-of-Thought}} 
& \multicolumn{2}{c}{\textbf{Counterfactual}} \\
\cmidrule(lr){2-3}\cmidrule(lr){4-5}\cmidrule(lr){6-7}\cmidrule(lr){8-9}
Model & AtAvg & QAvg & AtAvg & QAvg & AtAvg & QAvg & AtAvg & QAvg \\
\midrule
LLaMA-3.3-70B     & 0.369 & 3.640 & 0.370 & 3.658 & 0.370 & 3.652 & \boldblue{0.367} & 3.605 \\
Gemma-2-9B        & 0.362 & 3.837 & 0.361 & 3.837 & 0.362 & 3.831 & 0.354 & 3.722 \\
Gemma-2-27B       & 0.361 & 3.771 & 0.358 & 3.769 & 0.365 & 3.790 & 0.352 & 3.549 \\
Mistral-nemo      & 0.356 & 3.907 & 0.357 & 3.923 & 0.353 & \boldblue{3.907} & 0.348 & 3.705 \\
Mistral-small     & 0.372 & 3.718 & 0.365 & 3.708 & \boldblue{0.376} & 3.822 & 0.350 & \boldorange{3.322} \\
Olmo-2-13B        & 0.353 & \highest{3.933} & 0.350 & \boldblue{3.931} & 0.348 & 3.894 & 0.344 & 3.626 \\
Qwen-2.5-14B      & 0.362 & 3.783 & 0.362 & 3.776 & 0.368 & 3.829 & 0.356 & 3.572 \\
DeepSeek-LLaMA-70B& 0.353 & 3.837 & 0.353 & 3.876 & 0.351 & 3.840 & 0.353 & \boldblue{3.834} \\
DeepSeek-Qwen-14B & 0.370 & 3.872 & 0.369 & 3.881 & 0.369 & 3.882 & 0.366 & 3.828 \\
DeepSeek-Qwen-32B & \highest{0.376} & 3.763 & \boldblue{0.375} & 3.770 & 0.375 & 3.768 & 0.364 & 3.612 \\
\midrule
\midrule
\rowcolor{gray!20} Overall & 0.363 & 3.806 & 0.362 & \textbf{3.813} & 0.364 & \textbf{3.822} & 0.355 & 3.646 \\
\bottomrule
\end{tabular}
}
\vspace{-6mm}
\label{tab:reasoning_main}
\end{table}

\begin{wraptable}{r}{0.5\linewidth}
\caption{Average qualitative score (QAvg) and percentage of high-quality completions (Avg $>$ 4 across five rubric dimensions) for various prompt levels and reasoning paradigms, aggregated across 10 LLMs. Each cell reports Avg / High\%. Improvements are shown in green.}
\label{tab:summary}
\vspace{1mm}
\centering
\scriptsize
\begin{adjustbox}{width=\linewidth,center}
\begin{tabular}{lc|cc}
\toprule
\textbf{Prompt Specificity} & \textit{Avg QAvg / \%>4} & \textbf{Reasoning} & \textit{Avg QAvg / \%>4} \\
\midrule
Very low (L0) & 3.71 / \lowest{58.0} & Abductive & 3.81 / 63.4 \\
Low (L1) & 3.80 / 62.5 & Causal & 3.81 / 63.8 \\
Moderate (L2) & 3.80 / 62.8 & Chain-of-Thought & \highestnew{3.82} / \highestnew{64.0} \\
High (L3) & \highestnew{3.81} / 63.4 & Counterfactual & \lowest{3.64} / \lowest{54.9} \\
Very high (L4) & 3.76 / 58.4 &  &  \\
\bottomrule
\end{tabular}
\end{adjustbox}
\vspace{-3mm}

\end{wraptable}

\begin{figure}[!b]
\vspace{-5mm}
    \centering
    \includegraphics[width=0.6\linewidth]{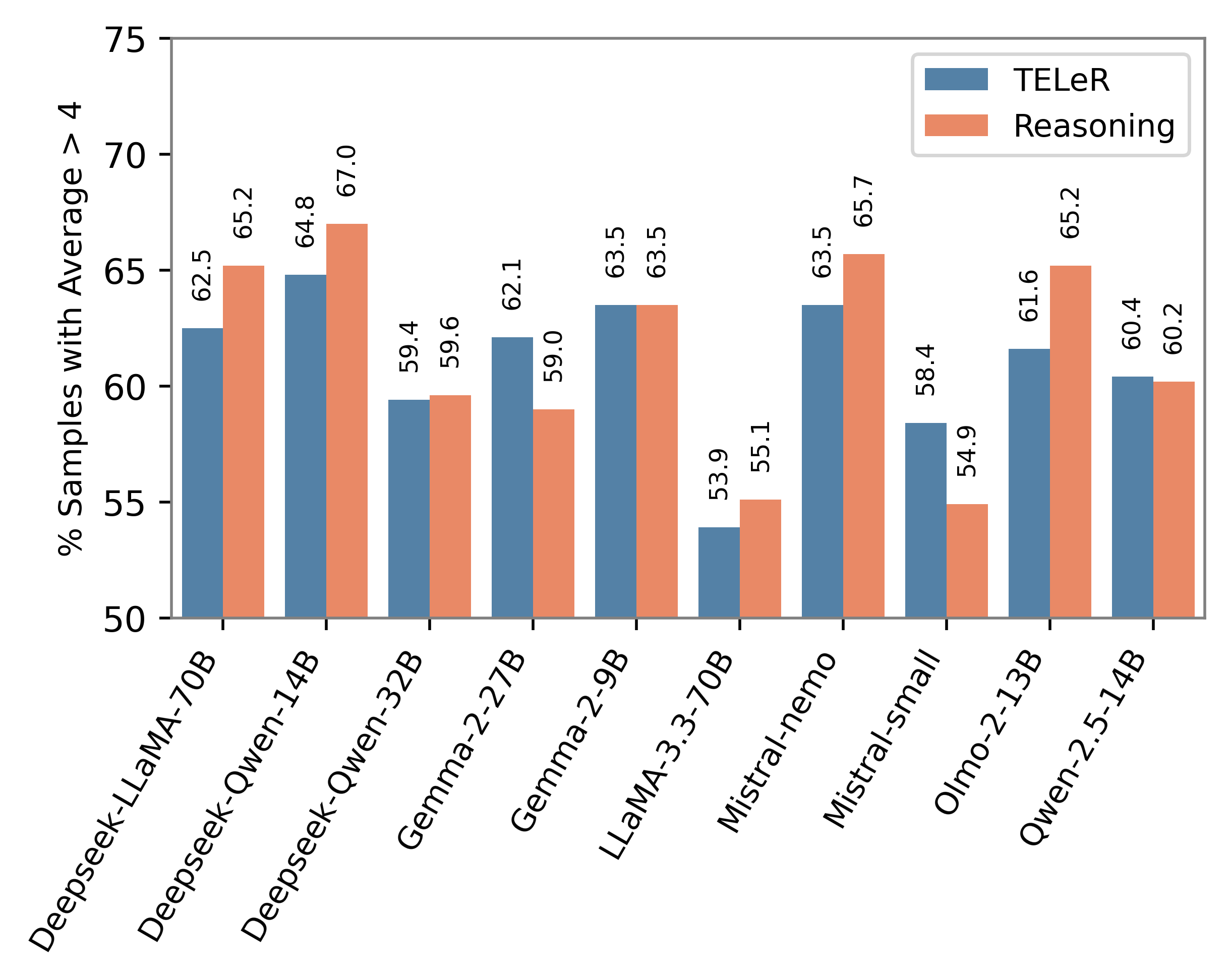}
    \caption{Percentage of high-quality completions (Avg $>$ 4 across five rubric dimensions) for each LLM under TELeR prompting and reasoning-based prompting. Scores are aggregated across all prompt levels and reasoning paradigms.}
    \label{fig:high-count}
    \vspace{-4mm}
\end{figure}

\noindent\textbf{Reasoning Provides Only Marginal Gains.}
Across the 10 evaluated models, reasoning yields only modest improvements over the L2 baseline. The average qualitative score increases slightly under abductive ($3.806$, $+0.2\%$), causal ($3.813$, $+0.3\%$), and chain-of-thought ($3.822$, $+0.6\%$) reasoning compared to the L2 baseline ($3.80$). At the individual model level, several models achieve their strongest performance under these paradigms. For example, \textit{Olmo-2-13B} attains the highest QAvg of $3.933$ under abductive reasoning and $3.931$ under causal reasoning.  However, the overall gains remain small, suggesting that larger models likely already engage in implicit reasoning during generation and that explicit reasoning prompts primarily refine rather than substantially alter their behavior.

\noindent\textbf{Counterfactual Reasoning Degrades Performance.}
In contrast, counterfactual thinking consistently reduces narrative quality: the average qualitative score drops to $3.64$, substantially lower than the other reasoning paradigms. This pattern suggests that encouraging alternative scenarios likely pushes the generated text away from the provided narrative context, weakening coherence and story consistency.



\noindent\textbf{Reasoning Improves High-Quality Infilling, But Not Uniformly.}
Table~\ref{tab:summary} reports the proportion of completions with a QAvg above 4, which we treat as an indicator of high-quality infilling. At the instruction level, High specificity achieves the highest rate ($63.4\%$), while very low and very high specificities trail behind. Among reasoning paradigms, Chain-of-Thought yields the highest proportion ($64.0\%$), whereas Counterfactual thinking produces substantially worse ($54.9\%$). Although the aggregated gains over the L2 baseline to reasoning are modest, Figure~\ref{fig:high-count} reveals substantial variation across individual models. Several models benefit noticeably from explicit reasoning guidance: Olmo-2-13B shows the largest improvement, increasing from $61.6\%$ to $65.2\%$ ($+3.6\%$), followed by DeepSeek-LLaMA-70B ($+2.7\%$), DeepSeek-Qwen-14B ($+2.2\%$), and Mistral-nemo ($+2.2\%$).  To identify the drivers of these differences, we examine per-dimension scores ($\geq 4$) across high-quality completions. We notice that Fluency, Bidirectional Coherence, Narrative Consistency, and Context Faithfulness all remain at or near ceiling ($\geq99.5\%$) across both settings, offering no discriminative signal. Only Informativeness varies meaningfully, rising from $95.9\%$ under instruction specificity to $97.2\%$ under reasoning ($+1.3\%$) consistently across all models. This suggests that reasoning guidance primarily helps models generate more content-rich completions that better fill the narrative gap, rather than improving surface fluency or narrative coherence. Qualitative examples of high-scoring completions are provided in Appendix~\ref{appendix:data-sample}.

\begin{table}[!b]
\vspace{-6mm}
\caption{\textbf{Top:} Average qualitative score and percentage of high-quality 
completions (Avg $>$ 4) by blank position across selective prompt specificity levels and reasoning paradigms. \textbf{Bottom:} Same breakdown by narratives from different domains. Each cell reports Avg / High\%. Improvements are shown in green.}
\vspace{1mm}
\label{tab:position-analysis}
\centering
\tiny
\setlength{\tabcolsep}{2pt}
\resizebox{0.48\linewidth}{!}{%
\begin{tabular}{lccc}
\toprule
\textbf{Setting} & \textbf{Opening} & \textbf{Middle} & \textbf{Closing} \\
\midrule
\multicolumn{4}{c}{\textit{Prompt Specificity}} \\
\midrule
Very Low  & 3.63 / 53.9\% & 3.82 / 63.0\% & 3.86 / 65.7\% \\
Moderate  & 3.80 / 63.2\% & 3.82 / 63.0\% & 3.78 / 60.8\% \\
Very High & 3.68 / 55.2\% & 3.80 / 61.9\% & 3.87 / 65.2\% \\
\midrule
\multicolumn{4}{c}{\textit{Reasoning Strategies}} \\
\midrule
Abductive        & 3.82 / 64.3\% & 3.81 / 63.0\% & 3.76 / 60.4\% \\
Chain-of-Thought & \highestnew{3.83 / 64.7\%} & 3.82 / \highestnew{63.8\%} & 3.79 / \highestnew{61.4\%} \\
Counterfactual   & 3.65 / 55.9\% & 3.64 / 54.7\% & 3.58 / 50.9\% \\
\midrule
\multicolumn{4}{c}{\textbf{Dataset-level Breakdown}} \\
\midrule
\multicolumn{4}{c}{\textit{Prompt Specificity (Aggregated)}} \\
\midrule
CNN/DailyMail & 3.91 / 67.9\% & 3.87 / 65.4\% & 3.88 / 65.3\% \\
Wikipedia     & 3.71 / 57.1\% & 3.78 / 60.5\% & 3.75 / 60.2\% \\
ROCStories    & 3.79 / 63.2\% & 3.87 / 67.0\% & 3.70 / 54.0\% \\
SIND          & 3.47 / 44.6\% & 3.53 / 48.6\% & 3.54 / 46.9\% \\
\midrule
\multicolumn{4}{c}{\textit{Reasoning Strategies (Aggregated)}} \\
\midrule
CNN/DailyMail & 3.83 / 63.8\% & 3.78 / 61.3\% & 3.77 / 60.8\% \\
Wikipedia     & 3.60 / 52.4\% & 3.71 / 57.3\% & 3.68 / 56.3\% \\
ROCStories    & \highestnew{3.88 / 69.3\%} & \highestnew{3.94 / 72.6\%} & \highestnew{3.79 / 55.0\%} \\
SIND          & \highestnew{3.59 / 50.8\%} & \highestnew{3.63 / 53.8\%} & \highestnew{3.62 / 48.8\%} \\
\bottomrule
\end{tabular}
}
\vspace{-2mm}
\end{table}

\subsection{Impact of Narrative Characteristics}
\noindent\textbf{Blank Position Difficulty Reverses Between Prompt Specificity and Reasoning.} Table~\ref{tab:position-analysis} (top) reveals that blank position interacts differently with the two instruction settings. Under prompt specificity, opening spans are most sensitive to instruction quality. For instance, the high-quality completion rate rises from $53.9\%$ at very low specificity to $63.2\%$ at moderate specificity (+$9.3\%$ improvement), while closing declines from $65.7\%$ to $60.8\%$. Meanwhile, under reasoning paradigms, this pattern reverses: opening improves further to $64.7\%$ under Chain-of-Thought ($+1.5\%$ over moderate specificity), while closing weakens to $61.4\%$ under Chain-of-Thought and drops to $50.9\%$ under counterfactual, the only condition where fewer than half of the completions are rated high quality. Across both settings, middle remains consistently stable ($63.0\%$--$63.8\%$), suggesting that having context on both sides buffers against the effects of instruction quality and reasoning strategy alike. The instability at opening and closing, by contrast, reflects a fundamental difference in what each position demands: opening spans require inferring what should logically come first, a task well suited to abductive reasoning, while closing spans require resolving an already established narrative where contextual fit matters more than reasoning patterns, explaining why the two positions respond differently to each guidance setting.

\noindent\textbf{Story-Based Narratives Benefit from Reasoning Across All Positions.} Table~\ref{tab:position-analysis} (bottom) shows that dataset difficulty is stable across blank positions in both settings, with CNN/DailyMail consistently the easiest and SIND the hardest regardless of where the gap appears. More importantly, the effect of reasoning guidance differs systematically by dataset type: ROCStories and SIND gain under reasoning across all three positions, with 
ROCStories showing the largest improvement: opening spans rise from $63.2\%$ to $69.3\%$ ($+6.1\%$) and middle from $67.0\%$ to $72.6\%$ ($+5.6\%$). SIND similarly improves across all positions: opening ($+6.2\%$), middle ($+5.2\%$), and closing ($+1.9\%$). In contrast, the high completion rate with CNN/DailyMail and Wikipedia declines consistently across all reasoning positions, with CNN/DailyMail dropping by $4.1$--$4.5\%$ and Wikipedia by $3.2$--$4.7\%$ across opening, middle, and closing spans. This suggests that reasoning guidance benefits story-based narratives where inference drives reconstruction, but offers less advantage for factual text where domain knowledge matters more than reasoning patterns.

\begin{table}[!htb]
\vspace{-4mm}
\caption{Percentage of average low-score ($<2$) cases by narrative length (\textit{\#sent}), blank position ($p$), and the number of sentences ($n$) to be filled. Percentages are computed across all LLMs for the best three prompt levels. The highest percentages are shown in orange.}
\vspace{1mm}
\centering
\small
\begin{adjustbox}{width=0.6\linewidth,center}
  \begin{tabular}{ccc|ccc} 
    \toprule
     \textit{\#sent.} & $(P,n)$  & low scores (\%) &  \textit{\#sent.} & $(P,n)$  & low scores (\%)\\
    \midrule
    5  & $(1,3)$   & \lowest[35]{16.0} & 21 & $(7,2)$  & 1.3  \\
    5  & $(1,2)$   & \lowest[30]{14.1}   &  30 & $(5,2)$   & 1.8 \\
    5  & $(2,2)$   & \lowest[25]{13.5}   &  30 & $(6,3)$   & 1.5  \\
    6  & $(1,1)$   & \lowest[20]{12.0}   & 33 & $(5,3)$  & 1.8  \\  
    5  & $(2,1)$   & \lowest[15]{11.2}  & 33 & $(4,2)$  & 0.8\\
    5  & $(3,1)$   & \lowest[10]{10.0}  & 38 & $(3,2) $  & 1.7  \\
    11 & $(3,3)$   & 1.7  & 45 & $(6,2)$   & 1.3\\
    13 & $(2,3)$   & 1.6  &  60 & $(8,3)$  & 1.2\\
    19 & $(4,1)$   & 1.0  & 60 & $(10,2)$  & 1.6 \\
    20 & $(7,3)$   & 1.9  & 62 & $(4,3)$   & 1.9 \\

    \bottomrule
  \end{tabular}
\end{adjustbox}

\label{tab:length_position_distribution}
\end{table}

\noindent\textbf{Short Narratives and Early Blanks Are the Worst-Case Combination.}
Further insight emerges when examining narrative length and blank position jointly. Table~\ref{tab:length_position_distribution} shows that low-scoring completions (average $<2$) occur predominantly in short narratives. Stories containing only 5–6 sentences account for most failure cases, with the highest low-score rates reaching $16.0\%$, $14.1\%$, and $13.5\%$ for configurations such as $(p=1, n=3)$, $(p=1, n=2)$, and $(p=2, n=2)$. In contrast, narratives with 20 or more sentences rarely produce poor outputs, with failure rates consistently below $2\%$ across positions and span lengths. 
This suggests that insufficient surrounding context, due to narrative brevity, is the primary driver of infilling failure, regardless of the blank position.

\section{Robustness Analyses}
Because our main results rely on stochastic generation and LLM-based evaluation, we conduct two additional analyses to examine the robustness of the reported findings to these sources of variation.

\paragraph{Generation Variability.}
We examine whether the reported trends are sensitive to stochastic decoding by repeating generation across five random seeds on a 10\% subset of the benchmark. We use four representative models: Gemma-2-2B, DeepSeek-Qwen-32B, Olmo-2-13B, and Mistral-Nemo under the best-performing instruction (L2) setting, with Olmo-2-13B and Mistral-Nemo additionally evaluated using Chain-of-Thought prompting. Across these settings, QAvg shows limited variation (SD = 0.012--0.044), suggesting that the observed trends are generally stable across different generations. 


\paragraph{Judge Sensitivity.}
We also examine whether the qualitative evaluation is sensitive to the choice of LLM judge. Using \texttt{Gemini 3.5 Flash-Lite} as an additional evaluator, we re-evaluate Gemma-2-2B and DeepSeek-Qwen-32B under Moderate (L2) instruction specificity, and Olmo-2-13B and Mistral-Nemo under Chain-of-Thought prompting. We find that the two judges produce nearly identical score gaps with the Moderate (L2) instruction setting, while the relative ordering of the two models with Chain-of-Thought prompting changes. Thus, while some comparisons remain consistent across evaluators, fine-grained model rankings can be sensitive to the choice of judge. Detailed results are provided in Appendix~\ref{app:robustness}.

\section{Discussion and Limitations}

\subsection{Benchmark Contamination Analysis}
\vspace{-1mm}
A natural concern with benchmarks drawn from public corpora is whether model performance reflects genuine infilling ability or familiarity with the pretraining data. We address this through two complementary analyses. First, we measured ROUGE similarity between model responses and ground-truth spans at a threshold of $0.8$ as a proxy for near-verbatim reproduction. We find that across all models and conditions, fewer than $0.15\%$ of responses exceeded this threshold ($1,624$ out of $1,279,880$ total responses), with the highest single-model rate at $0.30\%$ ($137$ out of $45,710$ for Gemma-2-27B), well under $1\%$ in every case. This finding is further corroborated by the consistently low lexical overlap scores reported in Appendix Table ~\ref{tab:fitb_full1} and Table~\ref{tab:fitb_full2}: if models were reproducing memorized content, we would expect substantially higher ROUGE scores than observed across all conditions. Beyond lexical evidence, the substantial variation in performance across models, prompt levels, and reasoning paradigms, including cases where smaller models outperform much larger ones, is more consistent with genuine capability differences than with uniform memorization effects. Taken together, these analyses suggest that direct reproduction of pretraining data is not a primary driver of our results, though we acknowledge that more rigorous methods such as membership inference~\citep{oren2023proving} would provide stronger guarantees and remain an important direction for future work.

\subsection{Limitations}
\label{sec:limitations}
\vspace{-1mm}
Our study has several limitations worth noting. First, the benchmark focuses exclusively on English narratives and instruction-tuned checkpoints, limiting generalization to other languages or model objectives. Second, the contiguous span masking strategy captures a common form of narrative gap but does not cover non-contiguous deletions or stylistic revisions that arise in real-world editing. The benchmark also focuses on narratives of moderate length and does not explicitly categorize masked spans by their semantic role or salience, which limits a more fine-grained analysis of what types of narrative information are most difficult to reconstruct. On the evaluation side, human annotations were collected on a subset of outputs and may miss rare failure modes, while LLM-based evaluation may not fully align with human judgment across all narrative dimensions. In addition, four of the five rubric dimensions approach ceiling performance in our high-quality analysis, which limits their discriminative power for distinguishing closely performing models. Finally, we use standardized decoding settings to ensure fairness across models, but this limits exploration of alternative decoding parameters that practitioners might tune in real-world applications. 


\section{Conclusion} 
\vspace{-1mm}
In this work, we introduced a narrative infilling benchmark of \textbf{9,142} instances to evaluate how well large language models can reconstruct missing story segments while preserving coherence with the broader narrative context. Through a systematic evaluation of 20 LLMs across varying instruction levels and reasoning strategies, we find that this capability remains challenging for current models with an average qualitative score of \textbf{3.80/5} and a maximum of \textbf{4.02/5}. Our results also show that instruction design plays an important role in infilling quality: moving from minimal to moderately detailed guidance improves performance by roughly \textbf{3\%}, while explicit reasoning guidance yields only marginal overall improvements (\textbf{<1\%}) but consistently increases the proportion of high-quality completions for story-based narratives. Beyond instruction effects, the narrative domain and blank position jointly shape task difficulty: short narratives pose the greatest challenge across all conditions, while position difficulty shifts depending on the type of guidance provided.

\section*{Acknowledgments}
This work has been partially supported by the National Science Foundation (NSF) Standard Grant Award \#2452028 and the Air Force Office of Scientific Research Grant/Cooperative Agreement Award \#FA9550-23-1-0426. We would also like to thank the University of Central Florida CS Department and the Institute of AI for their continuous support through Student Fellowships and Graduate Assistantships.


\bibliographystyle{plainnat}
\bibliography{custom}

\appendix
\section*{Appendix}
\label{sec:appendix}

\section{Additional Details}
\subsection{Response Extraction}
\label{sec:response_extraction}
A recurring challenge in narrative infilling with open-source LLMs is that their responses frequently contain extraneous material beyond the intended completion. For instance, models often preface with ``Here is the answer,'' restate the original prompt, or append unnecessary explanations to the actual fill. Therefore, extracting the true infilling span requires robust post-processing.

\noindent\textbf{Rule-Based Extraction. }
To address this, we developed a rule-based extraction pipeline by manually annotating 400 model outputs, identifying the correct answer spans, and recording recurring response structures. From this analysis, we derived a set of pattern-based extraction rules, summarized in Table~\ref{tbl:extract_patterns}. These rules handle a range of model-specific artifacts, including reasoning traces from DeepSeek models delimited by \texttt{</think>} tokens, redundant prompt restatements, and minor formatting inconsistencies across model families.

\noindent\textbf{Extraction Quality. }
We evaluated the extraction pipeline against the annotated set using ROUGE-1 to measure span-overlap quality. We achieve a ROUGE-1 score of \textbf{0.956}, a perfect match rate (PMR) of \textbf{91.5\%}, and a failure rate of only \textbf{3\%} (i.e., cases where ROUGE-1 is 0). These results indicate that our pipeline can reliably isolate the intended completion while discarding extraneous content, thereby providing clean, consistent inputs for downstream evaluation.

\begin{table}[htbp]
\caption{Common patterns in LLM responses and strategies to extract text from those responses.}
\centering
\resizebox{\textwidth}{!}{%
\begin{tabular}{ll}
\toprule
\textbf{Pattern} & \textbf{Extraction Strategy}\\
\midrule
\begin{tabular}{l}
DeepSeek models perform a reasoning step and terminate \\
the reasoned text with a \texttt{</think>} token.
\end{tabular}
  &
  \begin{tabular}{l}
Remove all the text that appears up to and including \texttt{</think>}.
  \end{tabular}\\ \hline
\begin{tabular}{l}
In many instances, models reproduce the problem text \\
with the generated answer filled in.
\end{tabular}
  & 
  \begin{tabular}{l}
Split text by the overlapping patterns between the problem \\
text and the generated answer.
  \end{tabular}\\ \hline
\begin{tabular}{l}
Many LLMs are adept at following formatting guidelines \\
given to them.
\end{tabular}
  & 
  \begin{tabular}{l}
Add additional formatting instructions and utilize manual \\
extraction patterns.
  \end{tabular}\\ \hline
\begin{tabular}{l}
Sometimes, LLMs slightly fail to adopt the proper formatting.
\end{tabular}
  & 
  \begin{tabular}{l}
Note the slight variations and add them to the extraction \\
patterns.
  \end{tabular}\\
\bottomrule
\end{tabular}
}

\label{tbl:extract_patterns}
\end{table}

\subsection{System-Level and Sample-Level Correlation}
\label{app:system-level-corr}
To analyze how well automatic evaluation metrics align with human judgments, we compute correlations between the scores assigned by the two evaluation methods. Following prior work on evaluation meta-analysis \citep{meta_sys_1, meta_sys_2, meta_sys_3, meta_sys_4}, correlations can be computed at two different levels: sample-level (completion-level) and system-level. We adopt the formulation from \citet{revisiting_gold}. Suppose $m$ models generate outputs for each of $n$ narrative infilling samples. Let $X$ and $Y$ denote the resulting $n \times m$ score matrices produced by two evaluation methods (human evaluation and an automatic metric).

\noindent\textbf{Sample-level correlation} measures how well the evaluation methods agree when comparing model outputs for each individual sample. It is computed by averaging the correlations across all samples:

$$
r_{sum}(X, Y) = \frac{1}{n} \sum_i \mathcal{C}(X_i, Y_i),
$$

where $X_i$ and $Y_i$ represent the scores assigned to the $m$ model outputs for the $i$-th sample, and $\mathcal{C}$ denotes a correlation function (e.g., Pearson correlation).

\noindent\textbf{System-level correlation} instead compares the overall rankings of models after aggregating scores across samples:

$$
r_{sys}(X, Y) = \mathcal{C}(\bar{X}, \bar{Y}),
$$

where $\bar{X}$ and $\bar{Y}$ contain the average scores of each model across the $n$ samples (e.g., $\bar{X}_j = \frac{1}{n}\sum_i X_{i,j}$).

\subsection{Computational Resources}
\label{sec:compute_resources}
All experiments were conducted on NVIDIA H100 GPUs (80GB HBM3). Models with up to 32B parameters were run on a single H100 GPU, while larger models (70B--72B parameters) required two H100 GPUs. Model inference was performed using the vLLM framework~\cite{kwon2023efficient} with models loaded from HuggingFace in bfloat16 precision. Given the large scale of our evaluation, 20 models across 5 prompt specificity levels and 4 reasoning paradigms on 9,142 instances, the total inference time varied by model size, ranging from approximately 8-10 hours for models under 7B parameters to 16-25 hours for 70B models per experimental condition, on a single or dual H100 GPU setup, respectively.




\begin{table*}[!htb]

\caption{Results across three specificity levels (L0 to L2). AtAvg is the mean of five automatic metrics. QAvg is the average across five infilling quality evaluation dimensions: Fluency, Context Faithfulness, Bidirectional Coherence, Narrative Consistency, and Informativeness. The green highlight indicates the highest score, while bold blue and orange indicate the highest and lowest average for each level. The highest across each metric is shown in bold. }
\centering
\tiny

\resizebox{0.85\textwidth}{!}{ 
\setlength{\tabcolsep}{2pt}
\begin{tabular}{lcccccc @{\hspace{8pt}}cccccc}
\toprule
Model 
& BERT & R-1 & Met & ChRF & F1 & \textbf{AtAvg} 
& Flu & Faith & BiCo & NarCo & Info & \textbf{QAvg} \\
\midrule

\multicolumn{13}{l}{\textbf{Very Low (L0)}} \\
\midrule
LLaMA-3.1-8B     & 0.849 & 0.139 & 0.119 & 0.147 & 0.301 & 0.311 & \textbf{4.364} & \textbf{3.777} & \textbf{3.737} & \textbf{3.973} & \textbf{3.540} & \boldblue{3.878} \\
LLaMA-3.2-1B         & 0.847 & 0.140 & 0.120 & 0.151 & 0.288 & 0.309 & 4.257 & 3.706 & 3.661 & 3.885 & 3.451 & 3.792 \\
LLaMA-3.2-3B         & 0.846 & 0.129 & 0.115 & 0.140 & 0.292 & 0.304 & 4.297 & 3.730 & 3.683 & 3.917 & 3.475 & 3.820 \\
LLaMA-3.3-70B        & 0.853 & 0.152 & 0.112 & 0.147 & 0.306 & 0.314 & 4.232 & 3.585 & 3.576 & 3.804 & 3.341 & 3.707 \\
Ministral-8B       & 0.845 & 0.130 & 0.105 & 0.130 & 0.288 & 0.300 & 4.270 & 3.698 & 3.651 & 3.887 & 3.432 & 3.788 \\
Mistral-nemo       & 0.843 & 0.120 & 0.086 & 0.120 & 0.266 & 0.287 & 4.261 & 3.653 & 3.632 & 3.857 & 3.366 & 3.754 \\
Mistral-small      & 0.849 & 0.144 & 0.108 & 0.149 & 0.287 & 0.307 & 4.218 & 3.521 & 3.515 & 3.759 & 3.249 & 3.652 \\
Gemma-2-2B          & 0.842 & 0.119 & 0.088 & 0.123 & 0.256 & 0.286 & 4.249 & 3.644 & 3.614 & 3.862 & 3.361 & 3.746 \\
Gemma-2-9B          & 0.850 & 0.137 & 0.097 & 0.132 & 0.286 & 0.300 & 4.275 & 3.650 & 3.627 & 3.875 & 3.375 & 3.760 \\
Gemma-2-27B         & 0.854 & 0.150 & 0.107 & 0.141 & 0.302 & 0.311 & 4.289 & 3.632 & 3.613 & 3.869 & 3.368 & 3.754 \\
Olmo-2-7B           & 0.842 & 0.120 & 0.085 & 0.117 & 0.263 & 0.285 & 4.164 & 3.578 & 3.560 & 3.799 & 3.302 & 3.680 \\
Olmo-2-13B          & 0.832 & 0.088 & 0.063 & 0.088 & 0.232 & 0.260 & 4.134 & 3.539 & 3.527 & 3.750 & 3.243 & 3.639 \\
Qwen-2.5-7B          & 0.853 & 0.144 & 0.104 & 0.151 & 0.300 & 0.310 & 4.299 & 3.707 & 3.697 & 3.926 & 3.482 & 3.822 \\
Qwen-2.5-14B         & 0.856 & 0.167 & \textbf{0.129} & \textbf{0.180} & 0.326 & \boldblue{0.332} & 4.295 & 3.696 & 3.676 & 3.909 & 3.514 & 3.818 \\
DeepSeek-LLaMA-8B  & 0.840 & 0.105 & 0.078 & 0.110 & 0.246 & 0.276 & 4.084 & 3.413 & 3.415 & 3.648 & 3.154 & 3.543 \\
DeepSeek-LLaMA-70B & 0.847 & 0.127 & 0.090 & 0.115 & 0.275 & 0.291 & 4.145 & 3.506 & 3.496 & 3.750 & 3.237 & 3.627 \\
DeepSeek-Qwen-1.5B & 0.836 & 0.097 & 0.073 & 0.103 & 0.232 & 0.268 & 3.748 & 3.095 & 3.112 & 3.340 & 2.889 & \boldorange{3.237} \\
DeepSeek-Qwen-7B   & 0.836 & 0.092 & 0.065 & 0.098 & 0.227 & \boldorange{0.264} & 4.024 & 3.374 & 3.362 & 3.621 & 3.101 & 3.496 \\
DeepSeek-Qwen-14B  & 0.855 & 0.152 & 0.115 & 0.154 & 0.312 & 0.318 & 4.267 & 3.638 & 3.623 & 3.877 & 3.413 & 3.764 \\
DeepSeek-Qwen-32B  & \textbf{0.860} & \textbf{0.167} & 0.120 & 0.157 & \textbf{0.331} & 0.327 & 4.237 & 3.522 & 3.524 & 3.778 & 3.300 & 3.672 \\

\midrule
\multicolumn{13}{l}{\textbf{Low (L1)}} \\
\midrule
LLaMA-3.1-8B         & 0.859 & 0.192 & 0.165 & 0.214 & 0.340 & 0.354 & 4.348 & 3.665 & 3.662 & 3.900 & 3.577 & 3.831 \\
LLaMA-3.2-1B         & 0.847 & 0.138 & 0.101 & 0.145 & 0.265 & \boldorange{0.299} & 4.106 & 3.386 & 3.391 & 3.624 & 3.164 & 3.534 \\
LLaMA-3.2-3B         & 0.856 & 0.183 & 0.166 & 0.223 & 0.319 & 0.349 & 4.218 & 3.357 & 3.380 & 3.633 & 3.326 & 3.583 \\
LLaMA-3.3-70B        & 0.863 & 0.214 & 0.193 & 0.240 & 0.367 & 0.375 & 4.220 & 3.451 & 3.460 & 3.707 & 3.371 & 3.642 \\
Ministral-8B       & 0.865 & 0.201 & 0.160 & 0.209 & 0.338 & 0.355 & 4.362 & 3.647 & 3.655 & 3.898 & 3.492 & 3.811 \\
Mistral-nemo       & 0.865 & 0.197 & 0.155 & 0.206 & 0.339 & 0.352 & 4.369 & 3.648 & 3.649 & 3.877 & 3.495 & 3.807 \\
Mistral-small      & 0.868 & \textbf{0.223} & \textbf{0.188} & \textbf{0.236} & 0.365 & \highest{0.376} & 4.278 & 3.504 & 3.526 & 3.763 & 3.400 & 3.694 \\
Gemma-2-2B          & 0.858 & 0.183 & 0.150 & 0.202 & 0.317 & 0.342 & \textbf{4.462} & \textbf{3.898} & \textbf{3.873} & \textbf{4.119} & \textbf{3.752} & \highest{4.021} \\
Gemma-2-9B          & 0.864 & 0.203 & 0.165 & 0.213 & 0.343 & 0.358 & 4.346 & 3.670 & 3.668 & 3.911 & 3.532 & 3.825 \\
Gemma-2-27B         & 0.868 & 0.214 & 0.174 & 0.218 & 0.353 & 0.365 & 4.348 & 3.630 & 3.631 & 3.871 & 3.488 & 3.794 \\
Olmo-2-7B           & 0.861 & 0.180 & 0.144 & 0.207 & 0.321 & 0.343 & 4.355 & 3.696 & 3.714 & 3.955 & 3.589 & 3.862 \\
Olmo-2-13B          & 0.862 & 0.185 & 0.152 & 0.207 & 0.324 & 0.346 & 4.342 & 3.626 & 3.645 & 3.884 & 3.513 & 3.802 \\
Qwen-2.5-7B          & 0.864 & 0.190 & 0.153 & 0.214 & 0.337 & 0.352 & 4.251 & 3.541 & 3.557 & 3.805 & 3.422 & 3.715 \\
Qwen-2.5-14B         & 0.865 & 0.207 & 0.173 & 0.227 & 0.352 & 0.365 & 4.301 & 3.621 & 3.632 & 3.857 & 3.494 & 3.781 \\
DeepSeek-LLaMA-8B  & 0.858 & 0.175 & 0.137 & 0.186 & 0.320 & 0.335 & 4.304 & 3.697 & 3.690 & 3.925 & 3.510 & 3.825 \\
DeepSeek-LLaMA-70B & 0.864 & 0.190 & 0.149 & 0.190 & 0.338 & 0.346 & 4.349 & 3.737 & 3.726 & 3.967 & 3.548 & 3.865 \\
DeepSeek-Qwen-1.5B & 0.851 & 0.151 & 0.121 & 0.165 & 0.287 & 0.315 & 3.853 & 3.118 & 3.144 & 3.379 & 2.969 & \boldorange{3.293} \\
DeepSeek-Qwen-7B   & 0.856 & 0.168 & 0.139 & 0.191 & 0.315 & 0.334 & 4.205 & 3.504 & 3.508 & 3.761 & 3.347 & 3.665 \\
DeepSeek-Qwen-14B  & 0.865 & 0.206 & 0.174 & 0.225 & 0.355 & 0.365 & 4.372 & 3.738 & 3.732 & 3.968 & 3.615 & 3.885 \\
DeepSeek-Qwen-32B  & \textbf{0.869} & 0.220 & 0.184 & 0.232 & \textbf{0.367} & 0.374 & 4.304 & 3.595 & 3.600 & 3.845 & 3.466 & 3.762 \\

\midrule
\multicolumn{13}{l}{\textbf{Moderate (L2)}} \\
\midrule
LLaMA-3.1-8B & 0.860 & 0.196 & 0.171 & 0.220 & 0.337 & 0.357 & 4.408 & 3.778 & 3.777 & 4.009 & \textbf{3.693} & 3.933 \\
LLaMA-3.2-1B  & 0.821 & 0.078 & 0.058 & 0.081 & 0.192 & \boldorange{0.246} & 4.089 & 3.495 & 3.472 & 3.698 & 3.282 & 3.607 \\
LLaMA-3.2-3B & 0.854 & 0.173 & 0.145 & 0.200 & 0.304 & 0.335 & 4.339 & 3.666 & 3.659 & 3.889 & 3.569 & 3.824 \\
LLaMA-3.3-70B & 0.863 & 0.214 & \textbf{0.194} & \textbf{0.239} & \textbf{0.365} & 0.375 & 4.229 & 3.453 & 3.469 & 3.707 & 3.378 & 3.647 \\
Ministral-8B  & 0.863 & 0.197 & 0.156 & 0.205 & 0.336 & 0.352 & 4.401 & 3.757 & 3.739 & 3.987 & 3.583 & 3.893 \\
Mistral-nemo & 0.864 & 0.201 & 0.167 & 0.215 & 0.340 & 0.357 & 4.384 & 3.683 & 3.695 & 3.912 & 3.555 & 3.846 \\
Mistral-small & 0.867 & 0.220 & 0.182 & 0.225 & 0.360 & 0.371 & 4.299 & 3.546 & 3.559 & 3.790 & 3.399 & 3.719 \\
Gemma-2-2B  & 0.857 & 0.177 & 0.146 & 0.194 & 0.313 & 0.337 & 4.442 & \textbf{3.880} & \textbf{3.838} & \textbf{4.092} & 3.689 & \boldblue{3.988} \\
Gemma-2-9B   & 0.863 & 0.202 & 0.167 & 0.213 & 0.344 & 0.358 & 4.370 & 3.711 & 3.702 & 3.950 & 3.577 & 3.862 \\
Gemma-2-27B  & 0.868 & 0.209 & 0.168 & 0.206 & 0.351 & 0.360 & 4.352 & 3.633 & 3.631 & 3.875 & 3.463 & 3.791 \\
Olmo-2-7B  & 0.860 & 0.168 & 0.127 & 0.183 & 0.310 & 0.329 & 4.389 & 3.759 & 3.769 & 3.991 & 3.612 & 3.904 \\
Olmo-2-13B  & 0.862 & 0.188 & 0.151 & 0.199 & 0.328 & 0.346 & \textbf{4.390} & 3.784 & 3.786 & 4.001 & 3.634 & 3.919 \\
Qwen-2.5-7B  & 0.864 & 0.188 & 0.152 & 0.212 & 0.336 & 0.350 & 4.281 & 3.597 & 3.615 & 3.848 & 3.471 & 3.762 \\
Qwen-2.5-14B  & 0.865 & 0.202 & 0.163 & 0.217 & 0.349 & 0.359 & 4.277 & 3.570 & 3.586 & 3.812 & 3.454 & 3.740 \\
DeepSeek-LLaMA-8B  & 0.859 & 0.182 & 0.146 & 0.199 & 0.320 & 0.341 & 4.305 & 3.681 & 3.680 & 3.924 & 3.524 & 3.823 \\
DeepSeek-LLaMA-70B & 0.864 & 0.195 & 0.160 & 0.205 & 0.339 & 0.353 & 4.341 & 3.721 & 3.721 & 3.959 & 3.566 & 3.862 \\
DeepSeek-Qwen-1.5B & 0.850 & 0.148 & 0.118 & 0.163 & 0.284 & 0.313 & 3.837 & 3.118 & 3.141 & 3.389 & 2.955 & \boldorange{3.288} \\
DeepSeek-Qwen-7B   & 0.855 & 0.166 & 0.137 & 0.191 & 0.307 & 0.331 & 4.184 & 3.514 & 3.513 & 3.769 & 3.348 & 3.666 \\
DeepSeek-Qwen-14B  & 0.866 & 0.211 & 0.181 & 0.232 & 0.356 & 0.369 & 4.363 & 3.728 & 3.723 & 3.971 & 3.607 & 3.878 \\
DeepSeek-Qwen-32B  & \textbf{0.869} & \textbf{0.221} & 0.185 & 0.234 & 0.365 & \boldblue{0.375} & 4.320 & 3.594 & 3.600 & 3.838 & 3.461 & 3.763 \\
\bottomrule
\end{tabular}
}
\label{tab:fitb_full1}
\end{table*}

\begin{table*}[t]

\caption{Results across two specificity levels (L4 and L5). AtAvg is the mean of five automatic metrics. QAvg is the average across five infilling quality evaluation dimensions: Fluency, Context Faithfulness, Bidirectional Coherence, Narrative Consistency, and Informativeness. The green highlight indicates the highest score, while bold blue and orange indicate the highest and lowest average for each level. The highest across each metric is shown in bold. }
\centering
\tiny

\resizebox{\textwidth}{!}{ 
\setlength{\tabcolsep}{2pt}
\begin{tabular}{lcccccc@{\hspace{8pt}} cccccc}
\toprule
Model 
& BERT & Rouge-1 & Metero & ChRF & Sem-F1 & \textbf{AtAvg} 
& Flu & Faith & BiCo & NarCo & Info & \textbf{QAvg} \\
\midrule

\multicolumn{13}{l}{\textbf{High (L3)}} \\
\midrule
LLaMA-3.1-8B  & 0.859 & 0.190 & 0.166 & 0.212 & 0.333 & 0.352 & 4.454 & 3.886 & \textbf{3.869} & 4.096 & \textbf{3.785} & \boldblue{4.018} \\
LLaMA-3.2-1B  & 0.828 & 0.087 & 0.064 & 0.094 & 0.208 & \boldorange{0.256} & 4.124 & 3.496 & 3.479 & 3.710 & 3.276 & 3.617 \\
LLaMA-3.2-3B  & 0.846 & 0.142 & 0.108 & 0.158 & 0.269 & 0.305 & 4.277 & 3.649 & 3.648 & 3.869 & 3.471 & 3.783 \\
LLaMA-3.3-70B & 0.863 & 0.211 & \textbf{0.187} & 0.232 & 0.360 & 0.371 & 4.229 & 3.456 & 3.477 & 3.714 & 3.361 & 3.647 \\
Ministral-8B  & 0.862 & 0.190 & 0.150 & 0.195 & 0.332 & 0.346 & 4.425 & 3.797 & 3.776 & 4.012 & 3.602 & 3.922 \\
Mistral-nemo  & 0.863 & 0.197 & 0.161 & 0.208 & 0.340 & 0.354 & 4.415 & 3.773 & 3.771 & 3.992 & 3.620 & 3.914 \\
Mistral-small & 0.866 & 0.214 & 0.176 & 0.219 & 0.352 & 0.365 & 4.293 & 3.566 & 3.568 & 3.810 & 3.414 & 3.730 \\
Gemma-2-2B  & 0.857 & 0.177 & 0.146 & 0.191 & 0.316 & 0.338 & \textbf{4.459} & \textbf{3.894} & 3.854 & \textbf{4.103} & 3.706 & 4.003 \\
Gemma-2-9B   & 0.863 & 0.202 & 0.167 & 0.212 & 0.343 & 0.357 & 4.375 & 3.724 & 3.722 & 3.958 & 3.584 & 3.873 \\
Gemma-2-27B  & 0.867 & 0.208 & 0.167 & 0.205 & 0.349 & 0.359 & 4.342 & 3.626 & 3.631 & 3.875 & 3.456 & 3.786 \\
Olmo-2-7B  & 0.860 & 0.172 & 0.132 & 0.189 & 0.317 & 0.334 & 4.387 & 3.762 & 3.772 & 4.001 & 3.608 & 3.906 \\
Olmo-2-13B  & 0.863 & 0.190 & 0.155 & 0.203 & 0.331 & 0.348 & 4.393 & 3.763 & 3.767 & 3.979 & 3.625 & 3.905 \\
Qwen-2.5-14B  & 0.864 & 0.202 & 0.163 & 0.218 & 0.349 & 0.359 & 4.278 & 3.592 & 3.604 & 3.833 & 3.461 & 3.754 \\
Qwen-2.5-7B  & 0.864 & 0.185 & 0.147 & 0.207 & 0.334 & 0.347 & 4.286 & 3.598 & 3.624 & 3.862 & 3.467 & 3.767 \\
DeepSeek-LLaMA-8B  & 0.855 & 0.166 & 0.127 & 0.179 & 0.304 & 0.326 & 4.260 & 3.629 & 3.633 & 3.860 & 3.433 & 3.763 \\
DeepSeek-LLaMA-70B & 0.863 & 0.195 & 0.159 & 0.204 & 0.338 & 0.352 & 4.345 & 3.716 & 3.710 & 3.966 & 3.558 & 3.859 \\
DeepSeek-Qwen-1.5B & 0.848 & 0.139 & 0.109 & 0.149 & 0.276 & 0.304 & 3.876 & 3.152 & 3.162 & 3.418 & 2.971 & \boldorange{3.316} \\
DeepSeek-Qwen-7B   & 0.853 & 0.159 & 0.130 & 0.182 & 0.300 & 0.325 & 4.156 & 3.450 & 3.460 & 3.710 & 3.277 & 3.611 \\
DeepSeek-Qwen-14B  & 0.866 & 0.211 & 0.181 & 0.232 & 0.357 & 0.369 & 4.365 & 3.737 & 3.737 & 3.972 & 3.615 & 3.885 \\
DeepSeek-Qwen-32B  & \textbf{0.869} & \textbf{0.221} & 0.185 & \textbf{0.234} & \textbf{0.366} & \boldblue{0.375} & 4.312 & 3.604 & 3.609 & 3.863 & 3.481 & 3.774 \\

\midrule
\multicolumn{13}{l}{\textbf{Very high (L4)}} \\
\midrule
LLaMA-3.1-8B & 0.847 & 0.146 & 0.105 & 0.149 & 0.283 & 0.306 & 4.217 & 3.690 & 3.658 & 3.868 & 3.453 & 3.777 \\
LLaMA-3.2-1B & 0.822 & 0.072 & 0.050 & 0.067 & 0.192 & 0.240 & 4.003 & 3.401 & 3.371 & 3.606 & 3.127 & 3.502 \\
LLaMA-3.2-3B  & 0.827 & 0.084 & 0.049 & 0.075 & 0.208 & 0.249 & 3.907 & 3.373 & 3.359 & 3.575 & 3.097 & 3.462 \\
LLaMA-3.3-70B  & 0.833 & 0.095 & 0.053 & 0.082 & 0.227 & 0.258 & 3.908 & 3.254 & 3.239 & 3.485 & 2.947 & 3.367 \\
Ministral-8B  & 0.852 & 0.150 & 0.110 & 0.143 & 0.301 & 0.311 & 4.365 & \textbf{3.826} & \textbf{3.789} & \textbf{4.014} & \textbf{3.562} & \boldblue{3.911} \\
Mistral-nemo   & 0.843 & 0.118 & 0.074 & 0.105 & 0.268 & 0.282 & 4.317 & 3.732 & 3.703 & 3.948 & 3.424 & 3.825 \\
Mistral-small & 0.866 & \textbf{0.205} & \textbf{0.159} & \textbf{0.199} & \textbf{0.354} & 0.357 & 4.346 & 3.669 & 3.673 & 3.901 & 3.481 & 3.814 \\
Gemma-2-2B   & 0.840 & 0.113 & 0.092 & 0.120 & 0.256 & 0.284 & 4.064 & 3.480 & 3.442 & 3.676 & 3.205 & 3.574 \\
Gemma-2-9B   & 0.857 & 0.168 & 0.124 & 0.157 & 0.317 & 0.324 & 4.330 & 3.725 & 3.694 & 3.947 & 3.475 & 3.834 \\
Gemma-2-27B   & 0.865 & 0.198 & 0.155 & 0.189 & 0.346 & 0.351 & \textbf{4.375} & 3.736 & 3.713 & 3.966 & 3.532 & 3.864 \\
Olmo-2-7B   & 0.845 & 0.127 & 0.095 & 0.133 & 0.273 & 0.295 & 4.336 & 3.788 & 3.761 & 3.988 & 3.533 & 3.881 \\
Olmo-2-13B  & 0.830 & 0.077 & 0.052 & 0.084 & 0.211 & 0.251 & 4.175 & 3.549 & 3.533 & 3.752 & 3.208 & 3.643 \\
Qwen-2.5-7B  & 0.853 & 0.137 & 0.099 & 0.148 & 0.295 & 0.306 & 4.223 & 3.559 & 3.559 & 3.811 & 3.320 & 3.695 \\
Qwen-2.5-14B  & 0.854 & 0.150 & 0.105 & 0.147 & 0.306 & 0.312 & 4.218 & 3.610 & 3.594 & 3.830 & 3.372 & 3.725 \\
DeepSeek-LLaMA-8B  & 0.832 & 0.091 & 0.056 & 0.092 & 0.211 & 0.257 & 3.877 & 3.280 & 3.282 & 3.495 & 3.035 & 3.394 \\
DeepSeek-LLaMA-70B & 0.859 & 0.180 & 0.144 & 0.183 & 0.318 & 0.337 & 4.264 & 3.650 & 3.631 & 3.882 & 3.456 & 3.777 \\
DeepSeek-Qwen-1.5B & 0.822 & 0.057 & 0.041 & 0.066 & 0.161 & \boldorange{0.229} & 2.647 & 2.226 & 2.280 & 2.407 & 2.082 & \boldorange{2.328} \\
DeepSeek-Qwen-7B   & 0.831 & 0.077 & 0.050 & 0.086 & 0.208 & 0.251 & 3.841 & 3.207 & 3.214 & 3.453 & 2.959 & 3.335 \\
DeepSeek-Qwen-14B  & 0.859 & 0.179 & 0.133 & 0.178 & 0.327 & 0.335 & 4.284 & 3.669 & 3.656 & 3.899 & 3.455 & 3.792 \\
DeepSeek-Qwen-32B  & \textbf{0.867} & 0.203 & 0.152 & 0.195 & 0.350 & \boldblue{0.353} & 4.272 & 3.559 & 3.560 & 3.813 & 3.373 & 3.715 \\

\bottomrule
\end{tabular}
}

\label{tab:fitb_full2}
\end{table*}

\begin{table*}[t]
\caption{Results across four reasoning paradigms at moderate specificity (L2). AtAvg is the mean of five automatic metrics. QAvg is the average across five infilling quality evaluation dimensions: Fluency, Context Faithfulness, Bidirectional Coherence, Narrative Consistency, and Informativeness. The green highlight indicates the highest score, while bold blue and orange indicate the highest and lowest average for each paradigm. The highest across each metric is shown in bold.}
\centering
\tiny
\setlength{\tabcolsep}{2pt}

\resizebox{\textwidth}{!}{
\begin{tabular}{lcccccc @{\hspace{8pt}}cccccc}
\toprule
Model 
& BERT & Rouge-1 & Metero & ChRF & Sem-F1 & \textbf{AtAvg} 
& Flu & Faith & BiCo & NarCo & Info & \textbf{QAvg} \\
\midrule

\multicolumn{13}{l}{\textbf{Abductive Reasoning}} \\
\midrule
LLaMA-3.3-70B      & 0.862 & 0.206 & 0.186 & 0.232 & 0.359 & 0.369 & 4.236 & 3.435 & 3.460 & 3.706 & 3.362 & 3.640 \\
Gemma-2-9B         & 0.864 & 0.206 & 0.172 & 0.222 & 0.346 & 0.362 & 4.347 & 3.682 & 3.682 & 3.915 & 3.558 & 3.837 \\
Gemma-2-27B        & 0.868 & 0.210 & 0.168 & 0.208 & 0.351 & 0.361 & 4.318 & 3.612 & 3.620 & 3.857 & 3.449 & 3.771 \\
Mistral-nemo       & 0.863 & 0.198 & 0.165 & 0.214 & 0.340 & 0.356 & 4.405 & 3.760 & 3.764 & 3.987 & 3.621 & 3.907 \\
Mistral-small      & 0.867 & 0.218 & 0.183 & 0.230 & 0.362 & 0.372 & 4.293 & 3.538 & 3.550 & 3.794 & 3.415 & 3.718 \\
Olmo-2-13B         & 0.863 & 0.192 & 0.161 & 0.213 & 0.335 & 0.353 & \textbf{4.403} & \textbf{3.785} & \textbf{3.789} & \textbf{4.017} & \textbf{3.672} & \highest{3.933} \\
Qwen-2.5-14B       & 0.865 & 0.202 & 0.167 & 0.225 & 0.351 & 0.362 & 4.300 & 3.623 & 3.624 & 3.863 & 3.506 & 3.783 \\
DeepSeek-LLaMA-70B & 0.864 & 0.196 & 0.160 & 0.203 & 0.342 & 0.353 & 4.331 & 3.697 & 3.695 & 3.927 & 3.535 & 3.837 \\
DeepSeek-Qwen-14B  & 0.866 & 0.211 & 0.181 & 0.232 & 0.360 & 0.370 & 4.350 & 3.719 & 3.734 & 3.957 & 3.598 & 3.872 \\
DeepSeek-Qwen-32B  & \textbf{0.870} & \textbf{0.222} & \textbf{0.185} & \textbf{0.235} & \textbf{0.369} & \boldblue{0.376} & 4.293 & 3.592 & 3.614 & 3.845 & 3.471 & 3.763 \\

\midrule
\multicolumn{13}{l}{\textbf{Causal Reasoning}} \\
\midrule
LLaMA-3.3-70B      & 0.862 & 0.207 & 0.185 & 0.234 & 0.360 & 0.370 & 4.229 & 3.465 & 3.481 & 3.732 & 3.385 & 3.658 \\
Gemma-2-27B        & 0.867 & 0.207 & 0.164 & 0.204 & 0.349 & 0.358 & 4.313 & 3.617 & 3.619 & 3.856 & 3.443 & 3.769 \\
Gemma-2-9B         & 0.863 & 0.205 & 0.171 & 0.220 & 0.345 & 0.361 & 4.340 & 3.681 & 3.692 & 3.917 & 3.556 & 3.837 \\
Mistral-nemo       & 0.863 & 0.198 & 0.167 & 0.218 & 0.342 & 0.357 & \textbf{4.403} & \textbf{3.781} & \textbf{3.782} & \textbf{4.005} & \textbf{3.644} & 3.923 \\
Mistral-small      & 0.865 & 0.209 & 0.174 & 0.227 & 0.350 & 0.365 & 4.269 & 3.534 & 3.539 & 3.792 & 3.408 & 3.708 \\
Olmo-2-13B         & 0.862 & 0.189 & 0.156 & 0.212 & 0.331 & 0.350 & 4.401 & 3.784 & 3.783 & 4.016 & 3.671 & \boldblue{3.931} \\
Qwen-2.5-14B       & 0.864 & 0.201 & 0.168 & 0.229 & 0.349 & 0.362 & 4.284 & 3.608 & 3.614 & 3.863 & 3.512 & 3.776 \\
DeepSeek-LLaMA-70B & 0.864 & 0.195 & 0.159 & 0.203 & 0.342 & 0.353 & 4.361 & 3.739 & 3.736 & 3.964 & 3.578 & 3.876 \\
DeepSeek-Qwen-14B  & 0.866 & 0.210 & 0.180 & 0.233 & 0.358 & 0.369 & 4.351 & 3.734 & 3.736 & 3.965 & 3.621 & 3.881 \\
DeepSeek-Qwen-32B  & \textbf{0.869} & \textbf{0.220} & \textbf{0.185} & \textbf{0.237} & \textbf{0.365} & \boldblue{0.375} & 4.297 & 3.599 & 3.617 & 3.852 & 3.483 & 3.770 \\

\midrule
\multicolumn{13}{l}{\textbf{Chain-of-Thought}} \\
\midrule
LLaMA-3.3-70B      & 0.862 & 0.207 & 0.188 & 0.233 & 0.361 & 0.370 & 4.236 & 3.457 & 3.470 & 3.722 & 3.376 & 3.652 \\
Gemma-2-27B        & 0.867 & 0.215 & 0.175 & 0.216 & 0.353 & 0.365 & 4.329 & 3.631 & 3.635 & 3.876 & 3.479 & 3.790 \\
Gemma-2-9B         & 0.864 & 0.207 & 0.173 & 0.222 & 0.347 & 0.362 & 4.345 & 3.674 & 3.676 & 3.907 & 3.552 & 3.831 \\
Mistral-nemo       & 0.863 & 0.194 & 0.161 & 0.209 & 0.339 & 0.353 & 4.399 & 3.769 & 3.763 & 3.996 & 3.610 & \boldblue{3.907} \\
Mistral-small      & \textbf{0.869} & \textbf{0.220} & \textbf{0.186} & \textbf{0.239} & \textbf{0.365} & \boldblue{0.376} & \textbf{4.334} & 3.653 & 3.673 & 3.910 & 3.540 & 3.822 \\
Olmo-2-13B         & 0.863 & 0.189 & 0.153 & 0.201 & 0.332 & 0.348 & 4.384 & \textbf{3.753} & \textbf{3.749} & \textbf{3.984} & \textbf{3.603} & 3.894 \\
Qwen-2.5-14B       & 0.865 & 0.209 & 0.177 & 0.233 & 0.357 & 0.368 & 4.327 & 3.671 & 3.673 & 3.906 & 3.567 & 3.829 \\
DeepSeek-LLaMA-70B & 0.863 & 0.193 & 0.158 & 0.202 & 0.340 & 0.351 & 4.325 & 3.698 & 3.706 & 3.928 & 3.542 & 3.840 \\
DeepSeek-Qwen-14B  & 0.866 & 0.210 & 0.180 & 0.231 & 0.359 & 0.369 & 4.359 & 3.732 & 3.737 & 3.973 & 3.611 & 3.882 \\
DeepSeek-Qwen-32B  & 0.869 & 0.221 & 0.184 & 0.231 & 0.368 & 0.375 & 4.299 & 3.602 & 3.616 & 3.849 & 3.471 & 3.768 \\

\midrule
\multicolumn{13}{l}{\textbf{Counterfactual Thinking}} \\
\midrule
LLaMA-3.3-70B      & 0.860 & 0.200 & 0.188 & 0.233 & 0.356 & \boldblue{0.367} & 4.215 & 3.394 & 3.414 & 3.665 & 3.337 & 3.605 \\
Gemma-2-27B        & 0.862 & 0.193 & 0.164 & 0.218 & 0.323 & 0.352 & 4.204 & 3.342 & 3.371 & 3.606 & 3.221 & 3.549 \\
Gemma-2-9B         & 0.860 & 0.192 & 0.166 & 0.221 & 0.328 & 0.354 & 4.294 & 3.532 & 3.557 & 3.793 & 3.434 & 3.722 \\
Mistral-nemo       & 0.859 & 0.186 & 0.160 & 0.215 & 0.321 & 0.348 & 4.272 & 3.519 & 3.545 & 3.773 & 3.415 & 3.705 \\
Mistral-small      & 0.855 & 0.182 & 0.170 & 0.232 & 0.313 & 0.350 & 4.038 & 3.068 & 3.112 & 3.368 & 3.023 & \boldorange{3.322} \\
Olmo-2-13B         & 0.858 & 0.177 & 0.154 & 0.215 & 0.315 & 0.344 & 4.242 & 3.419 & 3.452 & 3.673 & 3.342 & 3.626 \\
Qwen-2.5-14B       & 0.860 & 0.192 & 0.167 & 0.229 & 0.333 & 0.356 & 4.176 & 3.365 & 3.389 & 3.631 & 3.297 & 3.572 \\
DeepSeek-LLaMA-70B & \textbf{0.863} & \textbf{0.194} & \textbf{0.162} & \textbf{0.208} & \textbf{0.339} & 0.353 & \textbf{4.332} & \textbf{3.689} & \textbf{3.686} & \textbf{3.918} & \textbf{3.545} & \boldblue{3.834} \\
DeepSeek-Qwen-14B  & 0.864 & 0.205 & 0.178 & 0.235 & 0.349 & \boldblue{0.366} & 4.326 & 3.662 & 3.675 & 3.906 & 3.569 & 3.828 \\
DeepSeek-Qwen-32B  & 0.864 & 0.202 & 0.178 & 0.236 & 0.342 & 0.364 & 4.203 & 3.405 & 3.449 & 3.678 & 3.327 & 3.612 \\
\bottomrule
\end{tabular}
}

\label{tab:reasoning_full}
\end{table*}

\newpage

\subsection{Robustness Analyses}
\label{app:robustness}
Table~\ref{tab:seed_robustness} and ~\ref{tab:judge_sensitivity} report the generation-variability and LLM-judge-sensitivity analyses, respectively..


\begin{table}[!htb]
\centering
\caption{QAvg across five random seeds on a 10\% subset of the benchmark. Results are reported as mean $\pm$ standard deviation.}
\label{tab:seed_robustness}
\begin{tabular}{lcc}
\toprule
Model & Prompt Setting & QAvg $\pm$ SD \\
\midrule
Gemma-2-2B        & L2  & 4.2300 $\pm$ 0.0194 \\
DeepSeek-Qwen-32B & L2  & 4.0787 $\pm$ 0.0439 \\
OLMo-2-13B        & L2  & 4.1578 $\pm$ 0.0121 \\
Mistral-Nemo      & L2  & 4.1413 $\pm$ 0.0279 \\
OLMo-2-13B        & CoT & 4.1346 $\pm$ 0.0303 \\
Mistral-Nemo      & CoT & 4.2483 $\pm$ 0.0255 \\
\bottomrule
\end{tabular}
\end{table}

\begin{table}[!htb]
\centering
\caption{Judge sensitivity analysis on a 10\% subset of the benchmark. We report QAvg scores obtained using \texttt{GPT-4.1-mini} and \texttt{Gemini 3.5 Flash-Lite}.}
\label{tab:judge_sensitivity}
\begin{tabular}{llcc}
\toprule
Model & Setting & GPT-4.1-mini & Gemini 3.5 Flash-Lite \\
\midrule
Gemma-2-2B        & Moderate (L2) & 4.2496 & 4.1877 \\
DeepSeek-Qwen-32B & Moderate (L2) & 4.0256 & 3.9679 \\
OLMo-2-13B        & CoT           & 4.1527 & 4.3969 \\
Mistral-Nemo      & CoT           & 4.2837 & 4.2177 \\
\bottomrule
\end{tabular}
\end{table}

\section{Prompt Template}
\label{sec:prompt-template}


\begin{table*}[h]
\caption {Prompt template for five instruction specificity levels (very low to very high). }
\label{table:prompts}
\centering
\begin{tabular}{p{15cm}}

\begin{tcolorbox}[
    colback=gray!5,
    colframe=black!70,
    title=\textbf{System Message},
    fonttitle=\bfseries,
    sharp corners,
    boxrule=0.8pt
]
There is content that is missing from the given text. Fill in the blank.

\end{tcolorbox}


\begin{tcolorbox}[
    colback=gray!5,
    colframe=blue!60!black,
    title=\textbf{Prompt Level 0},
    fonttitle=\bfseries,
    sharp corners,
    boxrule=0.8pt
]
\begin{verbatim}
{{problem}}
\end{verbatim}

\end{tcolorbox}


\begin{tcolorbox}[
    colback=gray!5,
    colframe=blue!60!black,
    title=\textbf{Prompt Level 1},
    fonttitle=\bfseries,
    sharp corners,
    boxrule=0.8pt
]
\begin{verbatim}
{{problem}}
\end{verbatim}

Guess the missing {{n}} {{unit}} and make sure your answer begins with 
\texttt{<ANSWER>} and ends with \texttt{</ANSWER>}.

\end{tcolorbox}


\begin{tcolorbox}[
    colback=gray!5,
    colframe=blue!60!black,
    title=\textbf{Prompt Level 2},
    fonttitle=\bfseries,
    sharp corners,
    boxrule=0.8pt
]

\begin{verbatim}
{{problem}}
\end{verbatim}

The blank area is missing at \texttt{\{\{unit\}\}}. 
Fill in the blank with \texttt{\{\{n\}\}} sentences.

Surround your answer with the tags  
\texttt{<ANSWER>} and \texttt{</ANSWER>}.

Answer:

\end{tcolorbox}


\begin{tcolorbox}[
    colback=gray!5,
    colframe=blue!60!black,
    title=\textbf{Prompt Level 3},
    fonttitle=\bfseries,
    sharp corners,
    boxrule=0.8pt
]

\begin{verbatim}
{{problem}}
\end{verbatim}

\begin{itemize}[leftmargin=*,itemsep=0.2ex,partopsep=-0.6ex,parsep=-0.2ex]
\item Fill in the blank with \texttt{\{\{n\}\}} sentences.
\item The blank area is missing at \texttt{\{\{unit\}\}}.
\item Surround your answer with the tags \texttt{<ANSWER>} and \texttt{</ANSWER>}.
\end{itemize}

Answer:

\end{tcolorbox}

\begin{tcolorbox}[
    colback=gray!5,
    colframe=blue!60!black,
    title=\textbf{Prompt Level 4},
    fonttitle=\bfseries,
    sharp corners,
    boxrule=0.8pt
]

Example of a 1-word problem:  
"The quick brown \_\_\_\_\_ jumped over the lazy dog"

Answer: \texttt{<ANSWER>fox</ANSWER>}\\

Example of a 2-word problem:  
"The quick \_\_\_\_\_ jumped over the lazy dog"

Answer: \texttt{<ANSWER>brown fox</ANSWER>}

\begin{itemize}[leftmargin=*,itemsep=0.2ex,partopsep=-0.6ex,parsep=-0.2ex]
\item The blank area is missing \texttt{\{\{n\}\}} sentences at position \texttt{\{\{unit\}\}}.
\item Tell me the missing text.
\item Your response should follow the given examples.
\item Surround your answer with the tags \texttt{<ANSWER>} and \texttt{</ANSWER>}.
\end{itemize}

\begin{verbatim}
{{problem}}
\end{verbatim}

Answer:

\end{tcolorbox}

\end{tabular}

\end{table*}

\begin{table*}[h]
\caption{Prompts template for explicit reasoning guidance. }
\label{table:prompts}
\centering
\begin{tabular}{p{15cm}}

\begin{tcolorbox}[
    colback=gray!5,
    colframe=blue!60!black,
    title=\textbf{Instruction with Abductive Reasoning Guidance},
    fonttitle=\bfseries,
    sharp corners,
    boxrule=0.8pt
]

\textbf{Definition:}  
Abductive reasoning involves inferring the most plausible explanation for incomplete information. The model should select the text that best explains the surrounding narrative context.

\vspace{4pt}

\begin{verbatim}
{{problem}}
\end{verbatim}

The blank area is missing at \texttt{\{\{unit\}\}} of  \texttt{\{\{n\}\}} sentences. Use abductive reasoning to infer the most plausible missing text. Before generating, infer the most likely event that explains the missing part of the story given the surrounding context. Surround your answer with the tags \texttt{<ANSWER>} and \texttt{</ANSWER>}.

Answer:

\end{tcolorbox}


\begin{tcolorbox}[
    colback=gray!5,
    colframe=blue!60!black,
    title=\textbf{Instruction with Causal Reasoning guidance},
    fonttitle=\bfseries,
    sharp corners,
    boxrule=0.8pt
]

\textbf{Definition:}  
Causal reasoning focuses on understanding cause-and-effect relationships. The model should infer the missing text by considering events or conditions that logically lead to the surrounding narrative.

\vspace{4pt}

\begin{verbatim}
{{problem}}
\end{verbatim}

The blank area is missing at \texttt{\{\{unit\}\}} of  \texttt{\{\{n\}\}} sentences. Use causal inference to determine the missing text. First, identify the causal relationships in the surrounding narrative and then generate a sentence that maintains this logic. Surround your answer with the tags \texttt{<ANSWER>} and \texttt{</ANSWER>}. 

Answer:

\end{tcolorbox}

\begin{tcolorbox}[
    colback=gray!5,
    colframe=blue!60!black,
    title=\textbf{Instruction with Chain-of-Thought Reasoning },
    fonttitle=\bfseries,
    sharp corners,
    boxrule=0.8pt
]

\textbf{Definition:}  
Chain-of-thought reasoning encourages the model to reason step by step before arriving at the final answer, improving logical consistency and contextual understanding.

\vspace{4pt}

\begin{verbatim}
{{problem}}
\end{verbatim}

The blank area is missing at \texttt{\{\{unit\}\}} of  \texttt{\{\{n\}\}} sentences.  Fill in the blank by utilizing chain-of-thought reasoning. First, reason step-by-step about what likely happened in the story, then generate the missing text. Surround your answer with the tags \texttt{<ANSWER>} and \texttt{</ANSWER>}.

Answer:

\end{tcolorbox}

\begin{tcolorbox}[
    colback=gray!5,
    colframe=blue!60!black,
    title=\textbf{Instruction with Counterfactual Thinking},
    fonttitle=\bfseries,
    sharp corners,
    boxrule=0.8pt
]

\textbf{Definition:}  
Counterfactual reasoning considers alternative possibilities by asking what could have happened under different circumstances. The model should evaluate plausible alternatives and choose the text that best fits the narrative context.

\vspace{4pt}

\begin{verbatim}
{{problem}}
\end{verbatim}

The blank area is missing at \texttt{\{\{unit\}\}} of  \texttt{\{\{n\}\}} sentences.
Surround your answer with the tags \texttt{<ANSWER>} and \texttt{</ANSWER>}.
Use counterfactual thinking to fill in the blank. Consider possible alternative events consistent with the story and generate the most plausible missing part.

Answer:

\end{tcolorbox}
\end{tabular}
\end{table*}


\onecolumn
\begin{figure}[t]
\tiny
\begin{tcolorbox}[
width=\textwidth,
enhanced,
breakable,
colback=gray!5,
colframe=blue!60!black,
title=\textbf{Narrative Infilling Evaluation Prompt with \texttt{GPT-4.1-mini}},
fonttitle=\bfseries,
]
\small

\textbf{System Instruction}

You are an expert evaluator of narrative text infilling.  
Your job is to carefully analyze a story in which a portion is missing and assess how well a candidate's text fills that gap. This is a strict grading task.

\bigskip

\textbf{Evaluation Procedure} \smallskip

\textbf{STEP 1 — Read the story carefully.}

Identify: Characters \; | \; Established facts/events \; | \;Timeline and causal links \; | \; Tone and style

\textbf{STEP 2 — Insert the candidate infill.}

Replace ``\_\_\_\_'' with the candidate text and read the FULL story.

\textbf{STEP 3 — Check common failure modes (penalize strictly).}

Context copying/duplication (repeating sentences from the story) \; | \;
Generic filler (vague statements like ``things got worse'', ``he felt sad'')\; | \;
One-sided fit (fits left but not right, or vice versa)\; | \;
Unsupported additions (new facts/events/entities not implied)\; | \;
Narrative looping (story seems to happen twice)


\textbf{Scoring Rubric}\smallskip

\textbf{1. Fluency (1–5) — STRICT}

5 = Perfectly natural, polished, no issues \; | \; 4 = Strong, very minor awkwardness \; | \; 3 = Understandable but noticeably awkward or clunky \; | \;2 = Frequent awkward phrasing or grammar issues \; | \; 1 = Hard to read / broken language


\textbf{2. Context Faithfulness (1–5) — STRICT}

5 = Adds no contradictions and no unsupported claims; fully grounded \; | \;
4 = Mostly grounded; only tiny, reasonable assumptions\; | \;
3 = Some unsupported assumptions OR mild drift from context\; | \;
2 = Clear unsupported additions or tensions with context\; | \;
1 = Major contradiction or fabrication

\medskip

\textbf{3. Bidirectional Coherence (1–5) — STRICT}

(Must connect BOTH left and right; bridging matters.)

5 = Seamless bridge; next sentence feels inevitable\; | \;
4 = Strong bridge; minor gap but still smooth\; | \;
3 = Connects, but the transition is noticeably weak or incomplete\; | \;
2 = Weak / one-sided connection; feels abrupt\; | \;
1 = No meaningful connection to the story flow

\medskip

\textbf{4. Narrative Consistency (1–5) — STRICT}

(Story integrity: timeline, characters, tone, causality.)

5 = Fully consistent and story-appropriate in all aspects\; | \;
4 = Consistent overall; tiny inconsistency or tonal mismatch\; | \;
3 = Noticeable inconsistency OR structural disruption\; | \;
2 = Major inconsistency or strong structural disruption\; | \;
1 = Breaks the story world / makes the story nonsensical

\medskip

\textbf{5. Informativeness (1–5) — STRICT}

(Must actually fill the gap; penalize copying and generic filler.)

5 = Highly informative; provides the missing bridge clearly\; | \;
4 = Informative; fills the gap well with minor missing detail\; | \;
3 = Some useful content but incomplete or partially generic\; | \;
2 = Mostly generic, redundant, or minimal contribution\; | \;
1 = Adds nothing OR repeats/copies context


\textbf{Input}

\begin{verbatim}
STORY WITH BLANK:
{story_with_blank}

CANDIDATE INFILL:
{candidate_infill}
\end{verbatim}

\medskip

\textbf{Output Format (STRICT JSON)}

\begin{verbatim}
{
  "Fluency": 1-5,
  "Context_Faithfulness": 1-5,
  "Bidirectional_Coherence": 1-5,
  "Narrative_Consistency": 1-5,
  "Informativeness": 1-5
}
\end{verbatim}

\end{tcolorbox}
\end{figure}

\onecolumn
\section{Qualitative Samples}
\label{appendix:data-sample}
\tiny
\setlength{\tabcolsep}{6pt}
\renewcommand{\arraystretch}{1.1}

\begin{longtable}{p{0.12\textwidth} p{0.84\textwidth}}
\caption{Some samples of highly quality LLM-generated infills.}\\
\label{table:multi-domain-examples} \\
\toprule
\textbf{Field} & \textbf{Content} \\
\midrule
\endfirsthead

\toprule
\textbf{Field} & \textbf{Content (continued)} \\
\midrule
\endhead

\midrule
\multicolumn{2}{r}{\textit{Continued on next page}} \\
\endfoot

\endlastfoot


Problem &
Erin loved to play football. \_\_\_\_\_\_\_\_ He stopped playing football with them.\\

\addlinespace

Gold Answer &
He always played with his friends after school. His friends played too roughly for him. They often hurt him by tackling him.
\\
\addlinespace
LLM Response &
\NoSim{Unfortunately, Erin suffered a serious injury during a game, which made it impossible for him to continue playing.}
\\
\addlinespace

Scores &
Fluency: \textbf{5.0}  \; | \; Context Faithfulness: \textbf{5.0} \; | \; Bidirectional Coherence: \textbf{5.0} \; | \; Narrative Consistency: \textbf{5.0} \; | \; Informativeness: \textbf{5.0}
\\

\midrule

Problem &
One day, Quinn went to the library. \_\_\_\_\_\_\_\_ She returned the books five weeks later.\\

\addlinespace

Gold Answer &
Quinn picked up five books. She also got some copies of Vogue magazine. Then she checked out.
\\
\addlinespace
LLM Response &
\NoSim{She checked out several books on mystery novels. Excited to dive into the stories, she took them home and read one after another.} 
By the time she finished all of them, five weeks had passed.
\\
\addlinespace

Scores &
Fluency: \textbf{5.0}  \; | \; Context Faithfulness: \textbf{5.0} \; | \; Bidirectional Coherence: \textbf{4.0} \; | \; Narrative Consistency: \textbf{5.0} \; | \; Informativeness: \textbf{5.0}
\\

\midrule

Problem &
Tom was an avid tennis player. \_\_\_\_\_\_\_\_ He saved money by fixing his own racket string from then on.\\

\addlinespace

Gold Answer &
He frequently broke his strings. He was upset at paying large amounts of money to fix it. He bought his own racket stringer.
\\
\addlinespace
LLM Response &
\NoSim{One day, while playing, his racket string broke mid-match. Realizing the cost of getting it professionally restrung, he decided to learn how to do it himself.}
\\
\addlinespace

Scores &
Fluency: \textbf{5.0}  \; | \; Context Faithfulness: \textbf{5.0} \; | \; Bidirectional Coherence: \textbf{4.0} \; | \; Narrative Consistency: \textbf{5.0} \; | \; Informativeness: \textbf{5.0}
\\

\midrule

Problem &
Chester is a town in Crawford County, Arkansas, United States. It is part of the Fort Smith, Arkansas--Oklahoma Metropolitan Statistical Area. As of the 2020 Census, the population was 144. Chester was platted in 1887 when the railroad was extended to that point. A post office called Chester has been in operation since 1883. According to the United States Census Bureau, the town has a total area of 1.3 km$^2$ (0.5 mi$^2$), all land. As of the census of 2000, there were 99 people, 35 households, and 26 families residing in the town. The population density was 76.4/km$^2$ (197.1/mi$^2$). There were 46 housing units at an average density of 35.5/km$^2$ (91.6/mi$^2$). The racial makeup of the town was 89.90\% White and 10.10\% Native American. 2.02\% of the population were Hispanic or Latino of any race. There were 35 households, out of which 34.3\% had children under the age of 18 living with them, 48.6\% were married couples living together, 20.0\% had a female householder with no husband present, and 25.7\% were non-families. \_\_\_\_ The average household size was 2.83 and the average family size was 3.15. In the town, the population was spread out, with 31.3\% under the age of 18, 8.1\% from 18 to 24, 28.3\% from 25 to 44, 20.2\% from 45 to 64, and 12.1\% who were 65 years of age or older. The median age was 31 years. For every 100 females, there were 83.3 males. For every 100 females age 18 and over, there were 88.9 males. The median income for a household in the town was \$27,500, and the median income for a family was \$31,667. Males had a median income of \$26,250 versus \$14,583 for females. The per capita income for the town was \$11,269. There were 23.1\% of families and 38.9\% of the population living below the poverty line, including 48.6\% of those under eighteen and 25.0\% of those over 64.
\\

\addlinespace

Gold Answer &
14.3\% of all households were made up of individuals, and 2.9\% had someone living alone who was 65 years of age or older.
\\
\addlinespace
LLM Response &
\NoSim{28.6\% of all households were made up of individuals, and 11.4\% had someone living alone who was 65 years of age or older.}
\\
\addlinespace

Scores &
Fluency: \textbf{5.0}  \; | \; Context Faithfulness: \textbf{5.0} \; | \; Bidirectional Coherence: \textbf{5.0} \; | \; Narrative Consistency: \textbf{5.0} \; | \; Informativeness: \textbf{5.0}
\\

\midrule

Problem &
Jack wanted to grow vegetables. Unfortunately, he lived in an apartment. \_\_\_\_\_\_\_\_ Soon his herb garden grew.
\\

\addlinespace

Gold Answer &
He decided to grow herbs in a window box. Jack got the supplies and planted seeds.
\\
\addlinespace
LLM Response &
\NoSim{He didn't have a yard, but he had a sunny balcony. He decided to use containers and small pots to grow his herbs.}
\\
\addlinespace

Scores &
Fluency: \textbf{5.0}  \; | \; Context Faithfulness: \textbf{5.0} \; | \; Bidirectional Coherence: \textbf{5.0} \; | \; Narrative Consistency: \textbf{5.0} \; | \; Informativeness: \textbf{5.0}
\\

\midrule

Problem &
Chris and Joe were making eggs in a cast-iron skillet. Chris tried to wash the pan first with soap. \_\_\_\_\_\_\_\_ In the end the eggs tasted fine. Chris looked for an apology but got none.
\\

\addlinespace

Gold Answer &
Joe freaked out at him for ruining the seasoning of the pan.
\\
\addlinespace
LLM Response &
\NoSim{Joe scolded Chris for using soap, which can ruin the skillet's seasoning.}
\\
\addlinespace

Scores &
Fluency: \textbf{5.0}  \; | \; Context Faithfulness: \textbf{5.0} \; | \; Bidirectional Coherence: \textbf{4.0} \; | \; Narrative Consistency: \textbf{5.0} \; | \; Informativeness: \textbf{4.0}
\\

\bottomrule

\end{longtable}

\normalsize
\section{Human Evaluation Guidelines}
\label{appendix:annot-infill}

Annotators were provided with the following step-by-step instructions 
for evaluating each narrative infilling candidate.

\subsection*{Evaluation Procedure}

\paragraph{Step 1 — Read the story carefully.}
Before evaluating the candidate infill, read the full narrative and 
identify the following: (i) characters and their roles, (ii) established 
facts and events, (iii) timeline and causal links between events, and 
(iv) the overall tone and writing style.

\paragraph{Step 2 — Insert the candidate infill.}
Replace the blank (``\underline{\hspace{1cm}}'') with the candidate 
text and read the \textit{complete} story from beginning to end to 
assess how well the infill integrates with the surrounding narrative.

\paragraph{Step 3 — Check for common failure modes.}
Penalize strictly for any of the following:
\begin{itemize}[leftmargin=*]
    \item \textbf{Context copying or duplication:} repeating sentences 
    or phrases already present in the story.
    \item \textbf{Generic filler:} vague or uninformative statements 
    such as ``things got worse'' or ``he felt sad.''
    \item \textbf{One-sided fit:} the infill connects to the left 
    context but not the right, or vice versa.
    \item \textbf{Unsupported additions:} introducing new facts, 
    events, or entities not implied by the surrounding narrative.
    \item \textbf{Narrative looping:} the story appears to repeat 
    itself or restart due to the inserted text.
\end{itemize}

\subsection*{Scoring Rubric}

Each candidate infill is scored on five dimensions using a 1--5 
Likert scale. Annotators are instructed to apply each criterion 
strictly and independently.

\paragraph{1. Fluency (1--5).}
Measures the grammatical correctness and naturalness of the generated text.
\begin{itemize}[leftmargin=*]
    \item \textbf{5} = Perfectly natural and polished; no issues.
    \item \textbf{4} = Strong overall; very minor awkwardness.
    \item \textbf{3} = Understandable but noticeably awkward or clunky.
    \item \textbf{2} = Frequent awkward phrasing or grammatical issues.
    \item \textbf{1} = Hard to read or severely broken language.
\end{itemize}

\paragraph{2. Context Faithfulness (1--5).}
Measures how well the infill remains grounded in the surrounding 
narrative without introducing unsupported claims or contradictions.
\begin{itemize}[leftmargin=*]
    \item \textbf{5} = Fully grounded; adds no contradictions or 
    unsupported claims.
    \item \textbf{4} = Mostly grounded; only minor, reasonable 
    assumptions.
    \item \textbf{3} = Some unsupported assumptions or mild drift 
    from context.
    \item \textbf{2} = Clear unsupported additions or tensions with 
    the surrounding context.
    \item \textbf{1} = Major contradiction or fabrication.
\end{itemize}

\paragraph{3. Bidirectional Coherence (1--5).}
Measures how well the infill connects to \textit{both} the preceding 
and following context. Bridging both sides is required for a high score.
\begin{itemize}[leftmargin=*]
    \item \textbf{5} = Seamless bridge; the following sentence feels 
    inevitable.
    \item \textbf{4} = Strong bridge; minor gap but overall smooth.
    \item \textbf{3} = Connects to both sides but the transition is 
    noticeably weak or incomplete.
    \item \textbf{2} = Weak or one-sided connection; feels abrupt.
    \item \textbf{1} = No meaningful connection to the narrative flow.
\end{itemize}

\paragraph{4. Narrative Consistency (1--5).}
Measures the integrity of the infill with respect to the story's 
timeline, characters, tone, and causal structure.
\begin{itemize}[leftmargin=*]
    \item \textbf{5} = Fully consistent and story-appropriate in all 
    aspects.
    \item \textbf{4} = Consistent overall; tiny inconsistency or 
    minor tonal mismatch.
    \item \textbf{3} = Noticeable inconsistency or structural 
    disruption.
    \item \textbf{2} = Major inconsistency or strong structural 
    disruption.
    \item \textbf{1} = Breaks the story world or makes the narrative 
    nonsensical.
\end{itemize}

\paragraph{5. Informativeness (1--5).}
Measures how effectively the infill fills the narrative gap. 
Generic filler and context copying are penalized strictly.
\begin{itemize}[leftmargin=*]
    \item \textbf{5} = Highly informative; provides the missing 
    bridge clearly and specifically.
    \item \textbf{4} = Informative; fills the gap well with only 
    minor missing detail.
    \item \textbf{3} = Some useful content but incomplete or 
    partially generic.
    \item \textbf{2} = Mostly generic, redundant, or minimally 
    contributive.
    \item \textbf{1} = Adds nothing meaningful or directly repeats 
    the surrounding context.
\end{itemize}

\newpage
\section*{NeurIPS Paper Checklist}

\begin{enumerate}

\item {\bf Claims}
    \item[] Question: Do the main claims made in the abstract and introduction accurately reflect the paper's contributions and scope?
    \item[] Answer: \answerYes{} 
    \item[] Justification: Yes. The claims made in the abstract and introduction are supported by the results of our paper in the experiments section.

\item {\bf Limitations}
    \item[] Question: Does the paper discuss the limitations of the work performed by the authors?
    \item[] Answer: \answerYes{} 
    \item[] Justification: Yes. The limitations of the work are discussed in Section \ref{sec:limitations}.

\item {\bf Theory assumptions and proofs}
    \item[] Question: For each theoretical result, does the paper provide the full set of assumptions and a complete (and correct) proof?
    \item[] Answer: \answerNA{} 
    \item[] Justification: This work does not have any theoretical results.

    \item {\bf Experimental result reproducibility}
    \item[] Question: Does the paper fully disclose all the information needed to reproduce the main experimental results of the paper to the extent that it affects the main claims and/or conclusions of the paper (regardless of whether the code and data are provided or not)?
    \item[] Answer: \answerYes{} 
    \item[] Justification: All experimental details necessary for reproduction, including model configurations, decoding parameters, prompt templates, benchmark construction, and evaluation rubrics, are described in the main paper and Appendix~\ref{sec:prompt-template}. We additionally release the benchmark data, and evaluation code.

\item {\bf Open access to data and code}
    \item[] Question: Does the paper provide open access to the data and code, with sufficient instructions to faithfully reproduce the main experimental results, as described in the supplemental material?
    \item[] Answer: \answerYes{} 
    \item[] Justification: The dataset and evaluation code are publicly released. The repository URL is provided during submission.

\item {\bf Experimental setting/details}
    \item[] Question: Does the paper specify all the training and test details (e.g., data splits, hyperparameters, how they were chosen, type of optimizer) necessary to understand the results?
    \item[] Answer: \answerYes{} 
    \item[] Justification: Since our work involves evaluation rather than training, we report all relevant experimental details, including model information, decoding hyperparameters (temperature, top-p, maximum output length), prompt templates for all settings, and benchmark construction details. These are described in Section~\ref{sec:llms} and Appendix~\ref{sec:prompt-template}.

\item {\bf Experiment statistical significance}
    \item[] Question: Does the paper report error bars suitably and correctly defined or other appropriate information about the statistical significance of the experiments?
    \item[] Answer: \answerNo{} 
    \item[] Justification: Our experiments use deterministic or near-deterministic decoding settings applied uniformly across all models and conditions, minimizing run-to-run variability. Results are aggregated over large sample sizes ($9,142$ instances per model), providing stable estimates that do not require bootstrapping or repeated runs.

\item {\bf Experiments compute resources}
    \item[] Question: For each experiment, does the paper provide sufficient information on the computer resources (type of compute workers, memory, time of execution) needed to reproduce the experiments?
    \item[] Answer: \answerYes{} 
    \item[] Justification: The computational resources required for our experiments are stated in Appendix~\ref{sec:compute_resources}.
    
\item {\bf Code of ethics}
    \item[] Question: Does the research conducted in the paper conform, in every respect, with the NeurIPS Code of Ethics \url{https://neurips.cc/public/EthicsGuidelines}?
    \item[] Answer: \answerYes{} 
    \item[] Justification: The research conducted in the paper conforms with the NeurIPS Code of Ethics.

\item {\bf Broader impacts}
    \item[] Question: Does the paper discuss both potential positive societal impacts and negative societal impacts of the work performed?
    \item[] Answer: \answerNA{} 
    \item[] Justification: No immediate societal impacts of the work. 
    
\item {\bf Safeguards}
    \item[] Question: Does the paper describe safeguards that have been put in place for responsible release of data or models that have a high risk for misuse (e.g., pre-trained language models, image generators, or scraped datasets)?
    \item[] Answer: \answerNA{} 
    \item[] Justification: No such risks

\item {\bf Licenses for existing assets}
    \item[] Question: Are the creators or original owners of assets (e.g., code, data, models), used in the paper, properly credited and are the license and terms of use explicitly mentioned and properly respected?
    \item[] Answer: \answerYes{} 
    \item[] Justification: All works are cited.

\item {\bf New assets}
    \item[] Question: Are new assets introduced in the paper well documented and is the documentation provided alongside the assets?
    \item[] Answer: \answerYes{} 
    \item[] Justification: The submitted code and dataset for this submission are well documented within the corresponding \textsc{readme} file.

\item {\bf Crowdsourcing and research with human subjects}
    \item[] Question: For crowdsourcing experiments and research with human subjects, does the paper include the full text of instructions given to participants and screenshots, if applicable, as well as details about compensation (if any)? 
    \item[] Answer: \answerYes{} 
    \item[] Justification: Human evaluators used the same five-dimensional rubric applied to LLM-based evaluation, ensuring consistency between human and automatic assessment. The full evaluation criteria and scoring instructions provided to annotators are detailed in Appendix~\ref{appendix:annot-infill}. Annotators participated voluntarily as a part of the research project and received no monetary compensation.

\item {\bf Institutional review board (IRB) approvals or equivalent for research with human subjects}
    \item[] Question: Does the paper describe potential risks incurred by study participants, whether such risks were disclosed to the subjects, and whether Institutional Review Board (IRB) approvals (or an equivalent approval/review based on the requirements of your country or institution) were obtained?
    \item[] Answer: \answerNA{}. 
    \item[] Justification: It is not applicable in our experiments

\item {\bf Declaration of LLM usage}
    \item[] Question: Does the paper describe the usage of LLMs if it is an important, original, or non-standard component of the core methods in this research? Note that if the LLM is used only for writing, editing, or formatting purposes and does \emph{not} impact the core methodology, scientific rigor, or originality of the research, declaration is not required.
    \item[] Answer: \answerNA{} 
    \item[] Justification: LLM is used only for writing, editing, or formatting purposes.

\end{enumerate}





\end{document}